\documentclass{ws-dmaa}
\usepackage{CJKutf8}
\usepackage{xcolor}
\usepackage{url}
\usepackage{makecell}
\usepackage[biblabel,nomove]{cite}
\begin{document}
\markboth{Guanghua Wang, Weili Wu}
{Surveying Text Summarization with Deep Learning}
%
\catchline{}{}{}{}{}
%

\title{Surveying the Landscape of Text Summarization with Deep Learning: A Comprehensive Review}

\author{Guanghua Wang, Weili Wu}
\address{Computer Science Department, The University of Texas at Dallas, 800 W, Campbell Road\\
Richardson, 75080-3021, United States\, \\
\email{guanghua.wang@utdallas.edu\\
weiliwu@utdallas.edu}}

\maketitle
\begin{history}
\received{Day Month Year}
\revised{Day Month Year}
\accepted{Day Month Year}
\published{Day Month Year}
\end{history}

\begin{abstract}
In recent years, deep learning has revolutionized natural language processing (NLP) by enabling the development of models that can learn complex representations of language data, leading to significant improvements in performance across a wide range of NLP tasks. 
Deep learning models for NLP typically use large amounts of data to train deep neural networks, allowing them to learn the patterns and relationships in language data. This is in contrast to traditional NLP approaches, which rely on hand-engineered features and rules to perform NLP tasks.
The ability of deep neural networks to learn hierarchical representations of language data, handle variable-length input sequences, and perform well on large datasets makes them well-suited for NLP applications.
Driven by the exponential growth of textual data and the increasing demand for condensed, coherent, and informative summaries, text summarization has been a critical research area in the field of NLP. Applying deep learning to text summarization refers to the use of deep neural networks to perform text summarization tasks. 
In this survey, we begin with a review of fashionable text summarization tasks in recent years, including extractive, abstractive, multi-document, and so on. Next, we discuss most deep learning-based models and their experimental results on these tasks. The paper also covers datasets and data representation for summarization tasks. Finally, we delve into the opportunities and challenges associated with summarization tasks and their corresponding methodologies, aiming to inspire future research efforts to advance the field further.
A goal of our survey is to explain how these methods differ in their requirements as understanding them is essential for choosing a technique suited for a specific setting.
This survey aims to provide a comprehensive review of existing techniques, evaluation methodologies, and practical applications of automatic text summarization. 
\end{abstract}
\keywords{Natural language processing; Deep neural network; Text Summarization; Extractive Summarization; Abstractive Summarization; Multi-document Summarization}

\section{Introduction}
    \subsection{Overview of Deep Learning for Text Summarization}
    Text summarization is the process of reducing a text or multiple texts to their essential meaning or main points while preserving its overall meaning and tone \cite{luhn1958automatic}. It has a wide range of applications, from creating news headlines and abstracts to summarizing legal documents and scientific papers. One common application of summarization is in news aggregation, where summaries of news articles are provided to users to consume news content quickly and efficiently \cite{barzilay2005sentence, lee2005fuzzy}. Another important application of summarization is in the legal industry \cite{kanapala2019text, galgani2012combining}, where lawyers may need to quickly review large amounts of legal documents to identify relevant information. Summarization can also be used in healthcare to summarize medical records \cite{afantenos2005summarization, aramaki2009text2table}, which can help doctors and other healthcare professionals make more informed decisions. In the meanwhile, summarization is able to summarize social media posts for busy readers who want to stay informed but do not have time to read the entire document \cite{chua2013automatic, moussa2018survey}. \par
    
    Traditional summarization models typically include rule-based ones or statistical techniques that focus on identifying key phrases and sentences from the source text without relying on deep learning or complex language models. Traditional methods have been widely used and provide a foundation for understanding the summarization task. One of them is the keyword-based method, which focuses on identifying keywords within the text and using them to select or rank sentences. Common techniques contain Term Frequency-Inverse Document Frequency (TF-IDF) weighting \cite{christian2016single, lawrie2001finding}, where sentences with a high concentration of important keywords are considered more relevant. Another approach is the Heuristic method, which relies on predefined rules or heuristics, such as considering sentence position, length, or similarity to the title, to determine the important sentences. For example, the Lead method \cite{brandow1995automatic, wasson1998using} selects the first few sentences of a document, assuming that they contain the most critical information. Besides, graph-based systems represent the document as a graph, where nodes correspond to sentences and edges represent the relationships or similarities between them. Algorithms like PageRank \cite{mihalcea-tarau-2004-textrank, brin1998anatomy} or LexRank \cite{erkan2004lexrank} are used to identify the most important nodes (sentences) in the graph, which are then included in the summary. Latent Semantic Analysis (LSA) \cite{ozsoy2011text, steinberger2004using}, on the other hand, is a statistical method that aims to capture the underlying semantic structure of a document by reducing its dimensionality. LSA is applied to a term-sentence matrix, and Singular Value Decomposition (SVD) is used to identify the most significant concepts or topics. Sentences that best represent these concepts are selected for the summary. The SumBasic algorithm \cite{nenkova2005impact} calculates the probability of a word appearing in a summary based on its frequency in the document. Sentences are scored by averaging the probabilities of their words, and the highest-scoring sentences are chosen for the summary. This method is simple but can yield reasonably good results. \par
    However, the capacity of traditional methods to produce organized and smooth summaries or adjust to diverse fields or dialects is frequently deficient. These methods are often simpler and faster than modern approaches, but they may not be as effective or accurate in capturing the nuances of the source text. Most of them are mainly focused on extractive summarization, which involves selecting the most important and relevant sentences or phrases from the original document to create a concise summary. Due to the complexity of generating new text, traditional methods are less common in abstractive summarization, which aims to generate a condensed version of the source text by rephrasing and reorganizing the original content instead of merely extracting existing sentences.\par
    Nowadays, neural networks with multiple layers \cite{nallapati-etal-2016-abstractive, chopra-etal-2016-abstractive, see-etal-2017-get} enable the development of models that can understand, generate, and manipulate natural language from language data. Deep neural networks have demonstrated significant improvement in the performance of summarization tasks \cite{devlin-etal-2019-bert, lewis-etal-2020-bart}, especially when compared to traditional statistical and traditional machine learning approaches. Deep learning models can learn from large amounts of data and generate more accurate predictions by capturing complex patterns and relationships in language data \cite{radford2019language}. It can also handle the complexity and variability of natural language inputs, such as variable-length sequences of words and sentences. This allows the model to capture long-range dependencies and context, which are critical for understanding the meaning of a sentence or document. On the other hand, deep learning can also learn representations of language data end-to-end \cite{sutskever2014sequence}, without relying on hand-engineered features or rules. This approach enables the same model to be used for different tasks with minor modifications. With the large, general-purpose datasets \cite{nallapati-etal-2016-abstractive} and high-performance computation ability \cite{coates2013deep} in recent years, deep learning can use pre-trained models as a starting point for a new summarization task with limited annotated data. The rapid advancements in deep learning with new architectures and techniques have led to a steady stream of innovations in summarization, which has pushed the state-of-the-art in language understanding and generation.\par
    
    Numerous research papers have been published on the subject of text summarization in conjunction with deep learning. However, these papers vary in scope and focus: some primarily address prevalent models \cite{tas2007survey, liu2018graph, jangra2021survey, andhale2016overview, guan2020survey, bhatia2016automatic}, while others discuss the applications of summarization tasks \cite{abualigah2019text, gholamrezazadeh2009comprehensive, munot2014comparative, yadav2022automatic}. Some papers cover both aspects without delving into the datasets associated with text summarization \cite{nazari2019survey, gambhir2017recent}. Additionally, certain papers only review a specific sub-field of summarization \cite{allahyari2017text, nenkova2012survey, gupta2010survey, feng2021survey, moratanch2017survey, lin2019abstractive, koh2022empirical, wang2022survey}. This paper aims to provide a comprehensive overview of deep learning techniques for text summarization. This encompasses the key text summarization tasks and their historical context, widely adopted models, and beneficial techniques. Furthermore, a comparative analysis of performance across various models will be presented, followed by the prospects for their application. These aspects will be discussed in the subsequent sections. \par

    \subsection{Paper Structure}
    The rest of the paper is structured as follows:\par
    \begin{itemize}
    \item [$\blacktriangleright$] Section 2 presents a comprehensive review of different tasks and their brief histories in text summarization.
    \item [$\blacktriangleright$] Section 3 describes some of the most popular deep neural network models and related techniques.
    \item [$\blacktriangleright$] Section 4 reviews a recipe of datasets and shows how to quantify efficiency, and what factors to consider during the evaluation of each task.
    \item [$\blacktriangleright$] Section 5 discusses the main challenges and future directions for text summarization with deep learning techniques.
    \end{itemize}
    
\section{Tasks in Summarization}
    Summarization tasks can be classified into several categories based on different criteria, such as summarization method, source document quantity, source document length, summary length, and so on.
    
    \subsection{Summarization Method: Extractive vs Abstractive}
    Extractive and abstractive summarization are two primary approaches to text summarization. Extractive summarization aims to identify and select the most pertinent sentences or phrases from the original text \cite{mani1998machine}, whereas abstractive summarization creates novel sentences that rephrase and consolidate the key concepts of the source document \cite{dorr-etal-2003-hedge}. \par
    \begin{figure}[htbp]
      \centerline
      {\includegraphics[width=12cm]{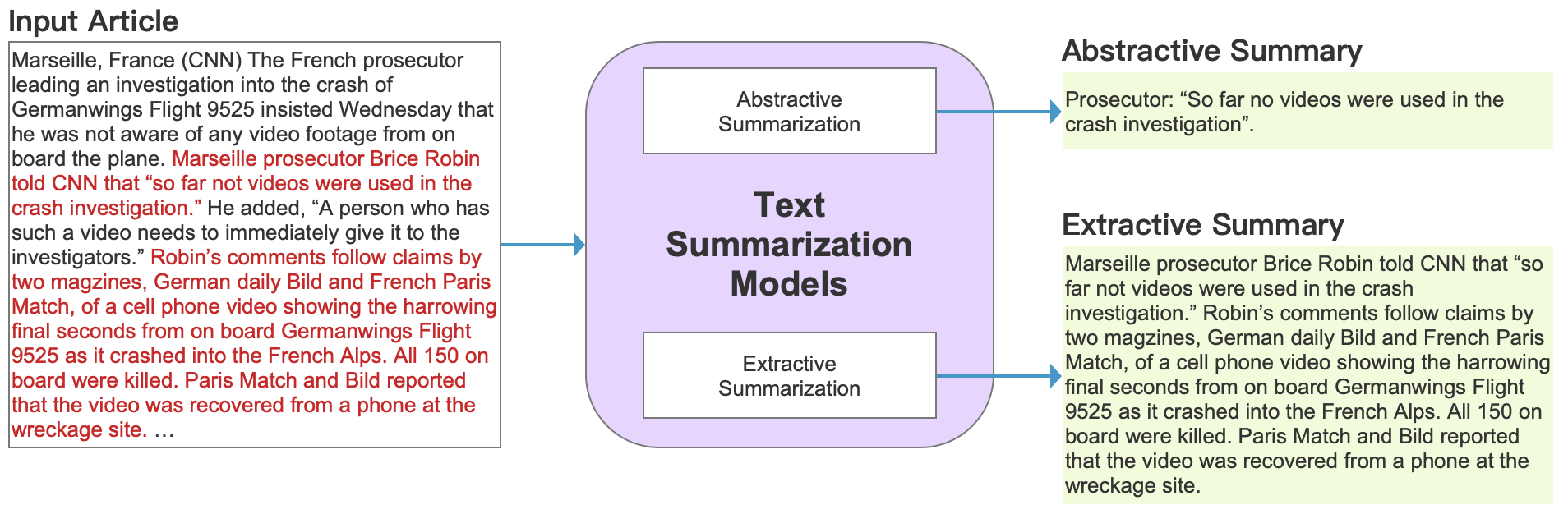}}
      \vspace*{8pt}
      \caption{The generated summary presented is the output of a "unilm-base-cased" model \cite{unilmv2} that has been fine-tuned, whereas the extractive summary provided is the output of a "distilbert-base-uncased" model \cite{sanh2020distilbert} that has also undergone fine-tuning. Both models were trained on the CNN/Daily Mail dataset. \cite{DaisyDeng2020}\label{fig1}}
    \end{figure}
    The underlying assumption of extractive summarization is that the original text contains sentences that are sufficiently informative and well-formed to be included in the summary directly. Extractive summarization typically relies on techniques such as scoring sentences based on their relevance, importance, or position within the source text. A graph-based ranking model for text processing, called TextRank \cite{mihalcea-tarau-2004-textrank}, is inspired by the PageRank algorithm used in web search. The model represents the input as a graph, where nodes are sentences or words, and edges represent the similarity between nodes. The algorithm iteratively scores nodes based on their connections, with higher-scoring nodes considered more important. Furthermore, LexRank \cite{erkan2004lexrank}, which is based on the concept of eigenvector centrality in a graph representation of the original text, is another unsupervised method for extractive summarization. On the other hand, Maximal Marginal Relevance (MMR) \cite{carbonell1998use} addresses the problem of redundancy of extractive summaries by selecting sentences that are both relevant to the query and diverse from the already-selected sentences in the summary. The MMR algorithm iteratively selects sentences based on the combination of query relevance and novelty, considering the content of previously chosen sentences.\par
    Instead of merely selecting existing sentences, abstractive summarization creates new sentences that convey the key ideas in a more natural and fluid manner. This approach requires a deeper understanding of the text and more advanced natural language generation capabilities. Goldstein et al. \cite{goldstein1999summarizing} propose a sentence extraction method based on a linear combination of several feature functions, followed by a sentence fusion step to generate abstractive summaries. Banko et al. \cite{banko2000headline} also use statistical models for content selection and surface realization to generate more succinct summaries. With probabilities calculated for candidate summary terms and their likely sequencing, a search algorithm is used to find a near-optimal summary in their system.\par
    While extractive summarization can produce coherent and accurate summaries, it may be limited in terms of fluency and flexibility, as the selected sentences are directly taken from the source text and may not always fit together seamlessly. Abstractive summarization with traditional methods can produce more creative and tailored summaries but faces challenges of flexibility and expressiveness. With the advent of deep learning, neural network-based models like sequence-to-sequence (Seq2Seq) models \cite{chopra-etal-2016-abstractive}, attention mechanisms \cite{rush-etal-2015-neural}, and transformers \cite{vaswani2017attention, lewis-etal-2020-bart} have significantly improved the performance of abstractive summarization by capturing complex patterns and semantic relationships within the original text, as well as extractive summarization. \par
    
    \subsection{Source Document Quantity: Single-document vs Multi-document}
    Single-document summarization and multi-document summarization are two distinct tasks within text summarization domain based on source document quantity. Single-document summarization focuses on generating a summary from a single input document, while multi-document summarization aims to create a summary by aggregating information from multiple related documents \cite{mckeown1995generating, radev2004centroid, dong2021two}. Within the single-document system, the objective is to condense the main ideas and essential information contained in that specific document. On the other hand, multi-document summarization tasks require identifying and combining the most relevant and non-redundant information from a set of documents, often covering the same topic or event. This means multi-document summarization has additional challenges, such as maintaining cross-document coherence, efficiently handling larger volumes of information, and processing redundancy across documents. These complexities make multi-document summarization generally more difficult than single-document summarization. \par
    Traditional approaches usually employ extractive techniques both in the context of single-document and multi-document summarization. Graph-based methods can be applied to both single-document and multi-document summarization by representing the relationships between sentences in one document or several documents as a graph, with sentences as nodes and the edges as the similarity between the sentences. The systems \cite{thakkar2010graph, lin2009graph, parveen2015topical, erkan2004lexrank} then use algorithms like PageRank, HITS, or LexRank to identify the most important sentences in the graph, which are then extracted and combined to form the summary. \par
    \begin{figure}[htbp]
      \centering
      \includegraphics[width=12cm]{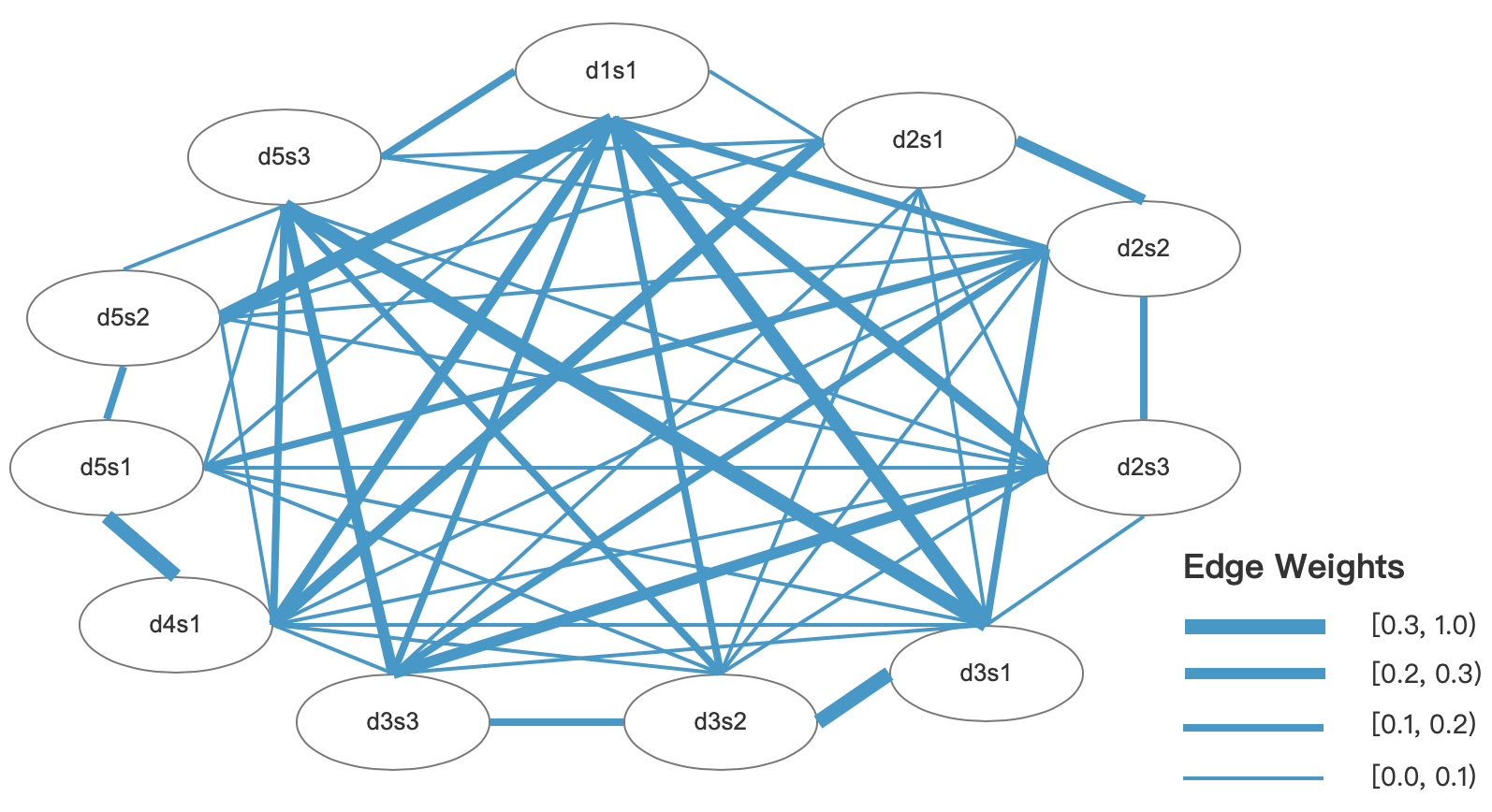}
      \vspace*{8pt}
      \caption{The weighted cosine similarity graph \cite{erkan2004lexrank} was generated for the cluster, based on the subset of d1003t from DUC 2004, which is a dataset for multi-document summarization tasks. The notation used in the figure is as follows: 'd' represents document and 's' represents sentence. For instance, d2s3 denotes the third sentence of document 2.}
    \end{figure}
    For single-document summarization, position-based methods \cite{edmundson1969new, kupiec1995trainable, lin2002manual} exploit the position of sentences within the document to identify important content. Additionally, the TF-IDF approach \cite{gong2001generic, mihalcea-tarau-2004-textrank, salton1988term} weighs the importance of words in the document based on their frequency and rarity. Sentences with high TF-IDF scores, which indicate a higher presence of significant words, are considered more important and are selected for the summary. Besides, Latent Semantic Analysis (LSA) \cite{steinberger2004using, gong2001generic, ozsoy2011text} is a mathematical technique that reduces the dimensionality of the term-document matrix and uncovers the underlying semantic structure of the document. By identifying the principal components or latent topics, LSA can rank sentences according to their relevance to these topics, and the top-ranked sentences are extracted for the summary. \par
    Regarding the task of multi-document summarization, Centroid-based methods \cite{radev2004centroid, erkan2004lexpagerank} calculate a central point for each document by considering the average term frequency of words in the document. Then centroids of all documents are used to compute an overall centroid, and sentences from different documents that are closest to this overall centroid are selected for the final summary. In the clustering \cite{wan2008multi, mckeown2002tracking, ganesan2010opinosis} approach, the documents are grouped into clusters based on their similarity to represent a common theme or topic. Sentences from each cluster are selected as representatives, based on features in the document. The final summary is generated by concatenating these representative sentences from each cluster. MMR \cite{carbonell1998use} is another technique that balances the relevance of the extracted information to the query and the diversity of the information to avoid redundancy in the summary of multi-document. \par
    Recently, deep learning techniques have made significant advancements in both single-document and multi-document summarization. They employ various architectures, such as RNNs \cite{nallapati2017summarunner, cao2015ranking}, Transformers \cite{zhang-etal-2019-hibert, liu-lapata-2019-hierarchical}, and pre-trained language models \cite{liu2018generative, xu2022sequence}, to generate coherent and informative summaries from single or multiple documents.
    
    \subsection{Source Document Length: Short vs Long}
    Short document summarization and long document summarization are two categories of text summarization, which differ based on the length and complexity of the input documents. Short document summarization focuses on generating summaries for relatively shorter documents, such as news articles \cite{nallapati-etal-2016-abstractive, grusky-etal-2018-newsroom}, blog posts \cite{hu2008comments}, or single research papers \cite{cachola-etal-2020-tldr}. Due to the limited length of the input, short document summarization often requires less context and background knowledge to produce coherent summaries. The main challenge is to effectively identify the most important information and convey it in a concise manner while maintaining coherence and readability. Meanwhile, long document summarization deals with generating summaries for more extensive and complex documents, such as books \cite{zhang2019generating, kryscinski2021booksum}, lengthy reports \cite{huang-etal-2021-efficient}, or collections of research papers \cite{cohan-etal-2018-discourse}. The primary challenge is to capture the overall theme and essential details while managing a large volume of information. This often needs advanced techniques to handle and process lengthy texts, maintain coherence, and produce summaries that effectively convey the critical points. Presently, a benchmark dataset whose source document length averages over 3,000 lexical tokens can be classified as long documents, given that most current state-of-the-art summarization systems (e.g., pre-trained models) are restricted to processing only 512 to 1,024 lexical tokens \cite{beltagy2020longformer}. These constraints cannot be easily overcome without new techniques that aid in enabling current architectures to process extensive textual inputs \cite{koh2022empirical}.\par
    Before the rise of deep learning, short document summarization primarily relied on algorithms like TF-IDF \cite{mihalcea-tarau-2004-textrank}, centroid \cite{erkan2004lexpagerank}, LSA \cite{steinberger2004using}, and graph-based models \cite{thakkar2010graph}, which are also used for the single-document summarization task discussed earlier. Due to limitations in system and algorithm capabilities, long document summarization was largely neglected until recent years. However, the emergence of deep neural networks has led to advancements in long document summarization task \cite{grail2021globalizing, cohan-etal-2018-discourse, xiao-carenini-2019-extractive, beltagy2020longformer}, as well as improvements to short document summarization \cite{devlin-etal-2019-bert, lewis-etal-2020-bart, sanh2020distilbert}.
    
    \subsection{Summary Length: Headline vs Short vs Long}
    The output summary length of summarization tasks can vary, depending on the desired level of detail and the intended use case. Based on summary length, there are three distinct text summarization tasks, which include headline summarization, short summary summarization, and long summary summarization.\par
    The aim of headline summarization is to generate a very brief and concise summary that captures the main theme or topic of the source text \cite{zhou2003headline, straka-etal-2018-sumeczech, zhou-hovy-2004-template, li2022news}. The output is usually a single sentence or a short phrase. Headline summarization is often used for news articles, where the goal is to provide readers with an immediate understanding of the main topic or event without diving into the details. This type of summarization task requires the model to extract or generate the most crucial information and convey it in a limited number of words. \par
    Short summary summarization \cite{DUC2004, zhang-etal-2021-emailsum, nallapati-etal-2016-abstractive} aims to produce a slightly longer summary that provides more context and details than a headline. Short summaries typically consist of a few sentences or a short paragraph. These summaries are useful for readers who want a quick overview of the source text without reading the entire document. Short summarization tasks require the model to identify and extract key points, main ideas, and essential information while maintaining the overall coherence and informativeness of the input text. \par
    The target of long summary summarization task \cite{zhang-etal-2021-emailsum, meng-etal-2021-bringing, makino-etal-2019-global} is to generate more comprehensive summaries that cover a wider range of topics, subtopics, or details from the source text. These summaries can be several paragraphs or even longer, depending on the length and complexity of the original document. This sort of task is suitable for situations where readers want to gain a deeper understanding of the source material without reading it in its entirety. It needs the model to not only extract key information but also maintain the logical structure and relationships between different ideas, making it a more challenging task. \par
    To produce concise summary headlines, Banko et al. \cite{banko2000headline} present a traditional approach that can generate summaries shorter than a sentence by building statistical models for content selection and surface realization. The approach is similar to statistical machine translation and uses statistical models of term selection and term ordering. Content selection and summary structure generation can be combined to rank possible summary candidates against each other using an overall weighting scheme. The Viterbi beam search \cite{huang2001guide} can be used to find a near-optimal summary.
    For short summary summarization, a trainable summarization program based on a statistical framework \cite{kupiec1995trainable} was developed, focusing on document extracts as a type of computed document summary. Features such as sentence length, fixed phrases, paragraph information, thematic words, and uppercase words are used to score each sentence. By employing a Bayesian classification function, the paper estimates the probability that a sentence will be included in a summary. \par
 
    In recent times, deep learning methods for headline summarization and short summary summarization, include sequence-to-sequence models, attention mechanisms \cite{10.5555/3171837.3171860}, and transformers like BERT \cite{devlin-etal-2019-bert}, GPT \cite{radford2018improving}, and T5 \cite{10.5555/3455716.3455856}, have emerged. These models are trained on large corpora to generate concise and informative headlines by learning the most important and relevant information within the input documents.
    Long summary summarization, along with long document summarization tasks, was overlooked for a significant period of time due to limitations in hardware and algorithm capabilities. Nevertheless, the development of deep neural networks has brought about significant progress in the field of long summary summarization techniques \cite{meng-etal-2021-bringing, makino-etal-2019-global}, like long document summarization tasks. \par
    
    \subsection{Language: Single-Language vs Multi-Language vs Cross-Lingual}
    Summarization tasks can be categorized based on the language involved, resulting in single-language, multi-language, and cross-lingual summarization.
    In single-language summarization, both the input documents and the generated summaries are in the same language. This is the most common type of summarization task, and the majority of research has focused on this area. 
    Multi-language summarization involves generating summaries for documents in various languages, but the output summary is in the same language as the input document. For instance, if the input is in French, the summary will be in French, if the input is in Japanese in the same model, the summary will be in Japanese \cite{patel2007language}.
    Cross-lingual summarization refers to the task of generating a summary in a target language for a source document written in a different language \cite{10.1145/979872.979877, orǎsan2008evaluation}. This type of summarization task requires models to not only understand and extract the main ideas from the source document but also translate the extracted information into the target language. \par
    \begin{CJK*}{UTF8}{gkai}
        \begin{center}
        \begin{table}
        \tbl{One example of cross-lingual summarization involves generating summaries of the same content in multiple languages \cite{bai-etal-2021-cross}.}{
        \begin{tabular}{cp{8cm}}
        \toprule
        Source text in English &  Crude oil futures climbed 2\% on Friday to a 28-month high, as the United States and Russia are in a deadlock over the Syrian issue, related concerns intensified. In October, the New York Mercantile Exchange's light sweet crude oil futures settlement price rose 2.16 US dollars to 110.53 US dollars per barrel. \\
        Source text in Chinese & \textit{原油期货周五攀升2\%，至28个月高点，因美俄两国在叙利亚问题上陷入僵局，相关担忧愈演愈烈。纽商所十月轻质低硫原油期货结算价涨2.16美元，至每桶110.53美元。} \\
        Summary in English & Oil prices hit a 28-month high as tensions in Syria escalated. \\
        Summary in Chinese & \textit{油价创28个月新高因叙利亚紧张局势升级。} \\
        \botrule
        \end{tabular}}
        \end{table}
        \end{center}
    \end{CJK*}
    Traditional methods such as keyword extraction \cite{radev-etal-2004-mead, gupta2013hybrid}, Hidden Markov Model \cite{fung2006one}, and graph-based algorithms \cite{azmi2012text} were commonly used for multi-language summarization before the widespread adoption of deep learning. These methods were adapted to work across various languages and often employed language-specific resources such as stop-words and stemming tools \cite{patel2007language} or language-agnostic features like TF-IDF \cite{gupta2013hybrid} to improve generalization.
    Cross-lingual summarization, on the other hand, typically depends on machine translation or bilingual dictionaries to comprehend and extract content from the source language, followed by producing a summary in the target language. Two pipeline methods were commonly employed: one approach is first summarizing the source document and then translating the summary to the target language \cite{10.1145/979872.979877, boudin2011graph, wan-2011-using, yao-etal-2015-phrase, zhang2016abstractive}, while the other approach entails translating the source document to the target language and then generating a summary \cite{orǎsan2008evaluation, wan-etal-2010-cross}. \par
    Nowadays, both multi-language and cross-lingual summarization tasks have benefited from advanced deep-learning techniques. These deep neural models \cite{chen-etal-2020-distilling, zheng2023long, ouyang-etal-2019-robust, duan-etal-2019-zero} can learn representations for multiple languages in a shared embedding space, which makes it possible to perform multi-language and cross-lingual transfer learning. By fine-tuning these models, researchers have achieved impressive results in generating summaries across different languages without the need for extensive parallel data or explicit translation. \par
    
    \subsection{Domain: General vs Specific domain}
    General summarization pertains to the process of generating summaries for any type of text or domain, without focusing on any particular subject matter. Conversely, specific domain summarization is designed to produce summaries for texts within a specific domain or subject matter, such as legal documents, scientific articles, or news stories. \par
    Specific domain summarization tasks typically demand domain-specific knowledge and may integrate specialized language models, ontologies, or rules to better capture the nuances and significant aspects of the target domain \cite{la-quatra-cagliero-2020-end, yu-etal-2021-adaptsum, 10.1093/bioinformatics/btz682, chalkidis-etal-2020-legal}. These systems may also consider the distinctive structure or format of texts in the domain, leading to better summarization results. Prior to deep learning, traditional specific domain summarization methods often incorporated domain-specific knowledge in the form of rules \cite{faeder2009rule, thione-etal-2004-hybrid}, or templates \cite{zhuang2006movie, wang-cardie-2013-domain, oya-etal-2014-template}. Additionally, feature-based methods were employed to identify important sentences or segments in the domain-specific summarization, such as the occurrence of certain keywords, phrases, or named entities relevant to the domain \cite{radev2001newsinessence, chen2006query, polsley-etal-2016-casesummarizer}. Graph-based algorithms, such as LexRank \cite{erkan2004lexrank} or TextRank \cite{mihalcea-tarau-2004-textrank}, were also commonly utilized for specific domain summarization. Domain-specific information could be incorporated into the graph representation or used to weigh the edges, making the algorithms more suited to the targeted domain. \par
    With the advent of deep learning, specific domain summarization has significantly evolved and improved. Domain adaptation is a technique that leverages pre-trained language models \cite{la-quatra-cagliero-2020-end, yu-etal-2021-adaptsum, zhang2021dsgpt}, which have been trained on vast amounts of text data and fine-tunes them for specific domain summarization tasks. To better capture the domain-specific knowledge, deep learning models also can be trained with domain-specific word embeddings or contextualized embeddings, such as BioBERT \cite{10.1093/bioinformatics/btz682} for the biomedical domain or Legalbert \cite{chalkidis-etal-2020-legal} for the Legal documents. These techniques offer several advantages over traditional approaches, including better representation learning, more effective handling of domain-specific knowledge, and the ability to adapt to new domains more easily. They have shown promising results in various specific domains and continue to push the boundaries of what is possible in domain-specific text summarization. \par
    
    \subsection{Level of Abstraction: Generic vs Query-focused}
    Query-focused summarization aims to generate a summary that addresses a specific user query or topic. The summary should contain the most relevant information from the source text with respect to the given query \cite{zhao2010query, chali2012query, abdullah-chali-2020-towards}. This type of summarization is useful for readers who have a specific question or interest and want a summary tailored to their needs. In contrast, the goal of generic summarization is to create a concise and coherent summary that captures the most important information from the source text. This type of summarization is not focused on any specific query or topic. Instead, it aims to provide an overview of the entire document or set of documents, which can be helpful for readers who want to quickly grasp the key points without going through the entire text. \par
    For query-focused, the relevance of the information is determined by its relation to the given query. Important information in a generic summary may be excluded if it is not directly relevant to the query, while less significant information related to the query may be included. Therefore, query-focused summarization typically requires additional processing to account for the given query. This may involve incorporating the query into the representation of the text, using query-specific features, or applying query-based attention mechanisms to guide the extraction or generation of the summary. These techniques are more specialized and require a query as input along with the source text. \par
    \begin{figure}[htbp]
      \centering
      \includegraphics[width=12cm]{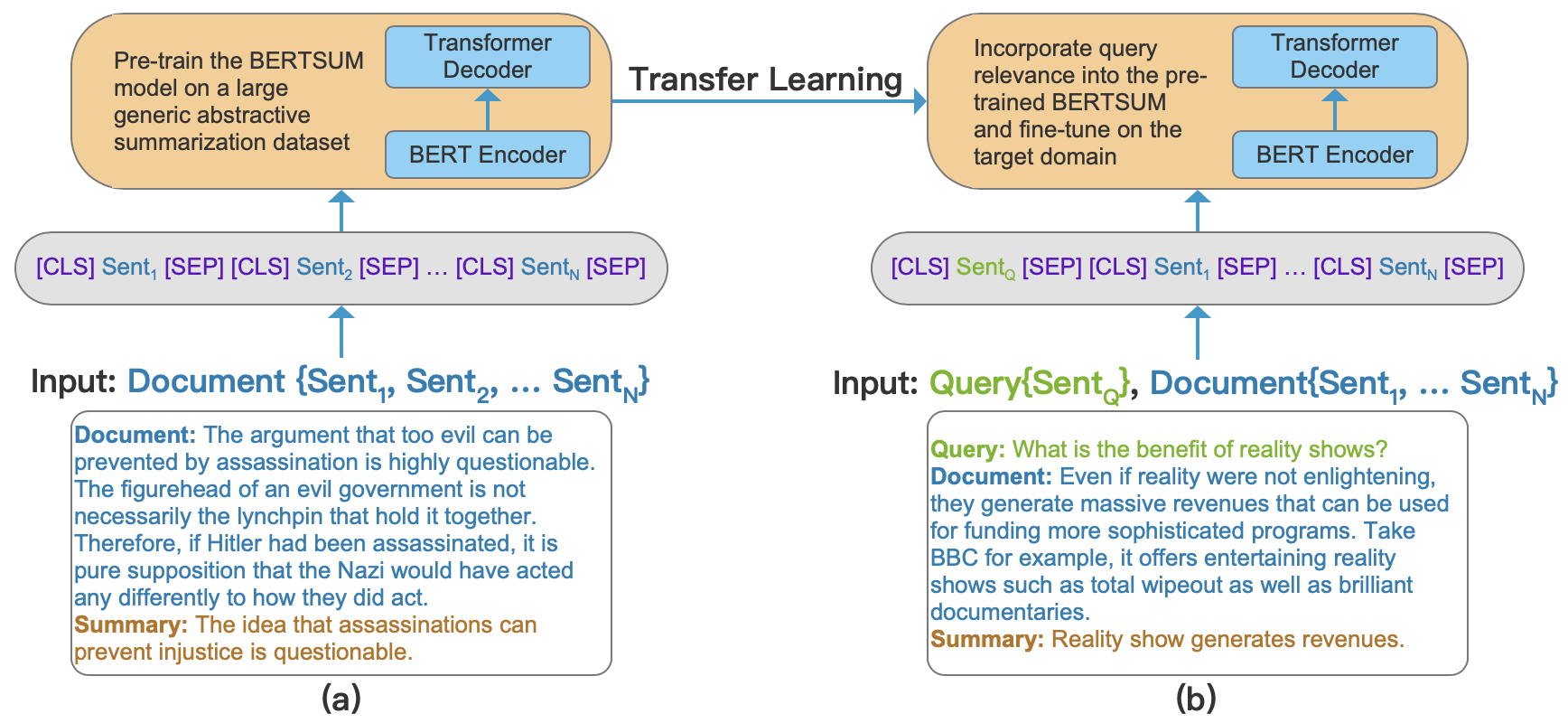}
      \vspace*{8pt}
      \caption{(a) Pre-training the BERTSUM \cite{liu2019fine} model on a generic abstractive summarization corpus, such as XSUM \cite{narayan-etal-2018-dont}. (b) Fine-tuning the pre-trained BERTSUM model on the target domain, which in this case is Debatepedia \cite{nema-etal-2017-diversity, laskar2020query}.}
    \end{figure}
    One of the simplest traditional approaches to query-focused summarization is to match the keywords in the query with those in the source text \cite{zhao2010query, daume-iii-marcu-2006-bayesian}. The sentences containing a higher number of matched keywords are considered more relevant and are included in the summary. Besides, query expansion methods \cite{chali2012query, chali-hasan-2012-effectiveness} expand the initial query using various techniques, such as synonym extraction or related term discovery, to improve the recall of relevant information. The expanded query is then used to match and rank sentences in the source text. This method helps to identify relevant information that may not have been captured by the original query. Furthermore, query-based weighting techniques incorporate the query by utilizing term weighting schemes like TF-IDF \cite{fisher2006query, daume-iii-marcu-2006-bayesian}. In these methods, query terms are given higher weights, which enhances the relevance score of sentences containing those terms. The sentences are then ranked based on their scores, and the top-ranked sentences are selected for the summary. Graph-based algorithms, like LexRank \cite{erkan2004lexrank} or TextRank \cite{mihalcea2004textrank}, can also be adapted for query-focused summarization by incorporating the query into the graph representation of the text. On the other hand, supervised machine learning like Support Vector Machines (SVM) \cite{fuentes-etal-2007-support, shen2011learning} or logistic regression \cite{10.1016/j.ipm.2010.03.005}, can be trained to rank the sentences in new, unseen documents based on their relevance to the query for query-focused summarization. \par
    With the appearance of deep learning techniques, such as Seq2Seq models \cite{vig-etal-2022-exploring}, pointer generator networks \cite{hasselqvist2017query}, pre-trained language models \cite{abdullah-chali-2020-towards}, and reinforcement learning \cite{cai2012mutually}, query-focused summarization could better understand the input and query context to generate more accurate and relevant summaries that address the specific user query or topic. \par

\section{Deep learning techniques}
    This section will review the most popular deep learning models and techniques utilized with deep neural networks, such as attention mechanisms, copy mechanisms, dictionary probabilities, etc. Each type of model has the potential to significantly improve various summarization tasks, which may be a blend of different categories, such as long abstractive legal document summarization, cross-lingual headline summarization, and extractive query-focused summarization.
    
    \subsection{Plain neural network}
    Plain neural network, also known as feed-forward neural network \cite{bengio2000neural, lecun2015deep}, is a type of artificial neural network that consists of several layers of interconnected nodes or neurons. The input is passed through one or more hidden layers, where the weights of each neuron are adjusted based on the error generated by the network's output. This process, called back-propagation, allows the network to learn how to classify or predict outputs based on the input data. \par
    Plain neural network is usually used in learning vector representation of words or sentences in summarization tasks. Word2vec models \cite{mikolov2013efficient, mikolov2013distributed}, which learns continuous vector representations of words from large amounts of text data, was developed with plain neural network. After these models, Glove \cite{pennington-etal-2014-glove} was proposed to combine the advantages of matrix factorization and local context window methods to create a more efficient and accurate model. The learned word vectors can capture various semantic and syntactic regularities, and can be used as features for different summarization tasks. On the other hand, Paragraph Vector \cite{hill-etal-2016-learning} learns fixed-length distributed representations of variable-length pieces of text by jointly predicting the words in the text and a separate paragraph-specific vector. \par
    \begin{figure}[htbp]
      \centering
      \includegraphics[width=12cm]{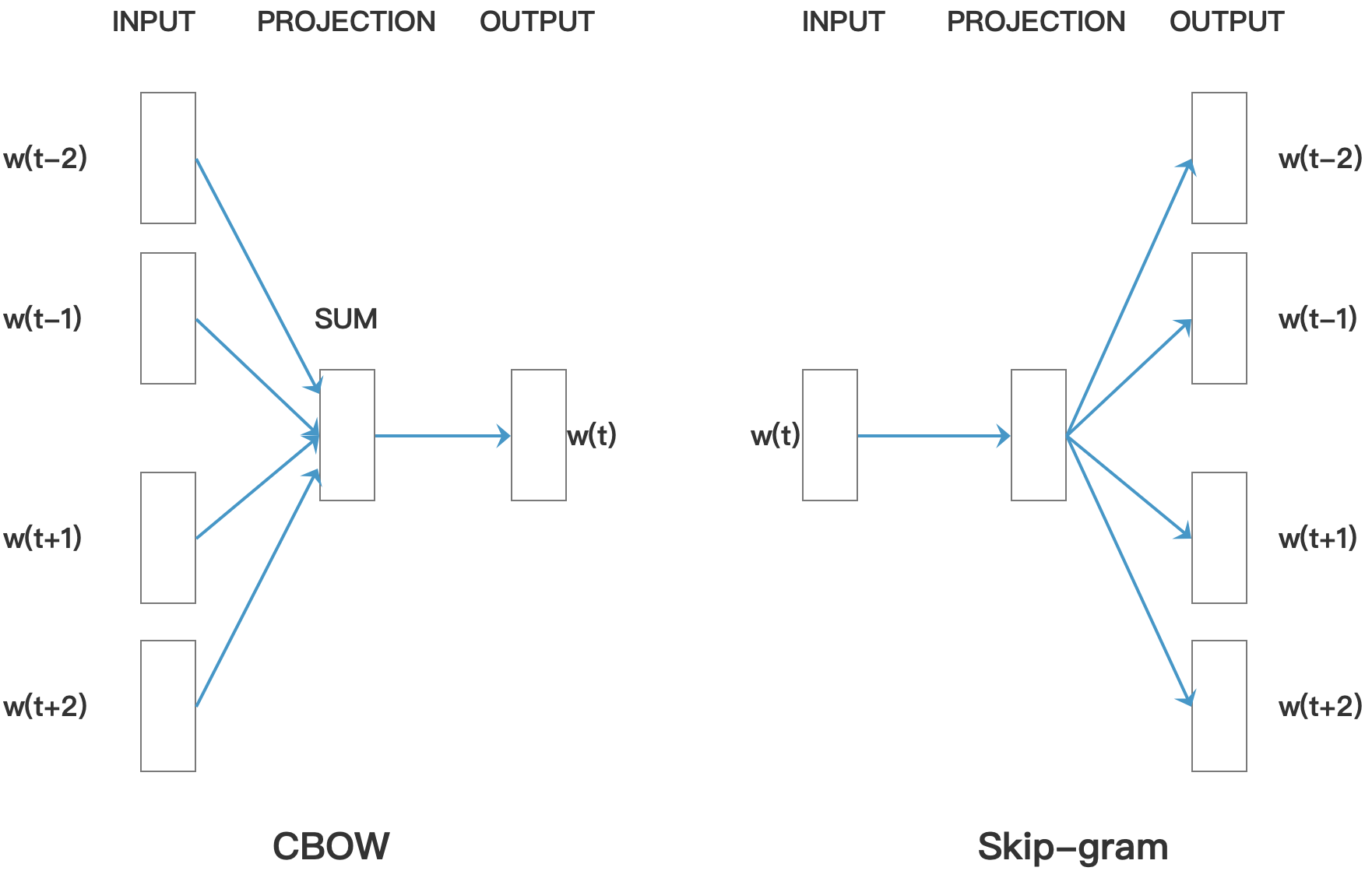}
      \vspace*{8pt}
      \caption{The current word was predicted based on the context by CBOW, and the surrounding words were predicted based on the current word by the Skip-gram.\cite{mikolov2013efficient}}
    \end{figure}
    
    \subsection{Recurrent neural network}
    Recurrent neural network (RNN) \cite{rumelhart1986learning, schuster1997bidirectional} is specifically designed to handle sequential data, making them highly suitable for NLP tasks. RNN can process variable-length sequences of inputs and maintain an internal state, which allows the model to remember information from previous time steps. However, RNN is prone to vanishing and exploding gradient problems, which makes it challenging to learn long-range dependencies. \par
    To address these issues, Long Short-Term Memory (LSTM) \cite{hochreiter1997long} and Gated Recurrent Unit (GRU) \cite{cho-etal-2014-learning} networks were developed to selectively remember or forget information over time. Each memory cell in an LSTM network has three main components: an input gate, a forget gate, and an output gate. The input gate determines how much of the new input information should be stored in the memory cell, while the forget gate determines how much of the previous memory state should be forgotten. The output gate controls how much of the current memory state should be used to generate the output. As opposed to LSTM, GRU uses a simpler gating mechanism that has only two gates: an update gate controls how much of the previous memory state should be retained and how much of the new input should be added to the memory cell, a reset gate determines how much of previous memory state should be ignored in favor of the new input. \par
    For extractive multi-document summarization, SummaRuNNer \cite{nallapati2017summarunner} is a GRU-based RNN model, which allows visualization of its predictions based on abstract features such as information content, salience, and novelty. It is an extractive model using abstractive training, which can train on human-generated reference summaries alone, removing the need for sentence-level extractive labels. \par
    The attention mechanism is a technique that allows models to focus on different parts of the input when producing an output. Both Chopra et al. \cite{chopra-etal-2016-abstractive} and Nallapati et al. \cite{nallapati-etal-2016-abstractive} have investigated the utilization of Attentional RNN Decoder for enhancing abstractive summarization performance. In the former study, a Convolutional attention-based network is employed as the encoder, furnishing a conditioning input to guide the decoder in focusing on relevant portions of the input sentence during word generation. On the other hand, the latter study employs an RNN-based encoder and incorporates keyword modeling to capture the hierarchical structure between sentences and words, effectively addressing the challenge of rare or unseen words. To tackle this issue, they propose a novel switching decoder/pointer architecture that enables the model to make decisions between generating a word and indicating its location within the source document. \par
    Furthermore, See et al. \cite{see-etal-2017-get} presents a hybrid pointer network that copies words from the source text to reproduce accurate information and handle out-of-vocabulary words. They also developed a coverage architecture to avoid repetition in the summary. This aids the attention mechanism to avoid repeatedly attending to the same locations, reducing the generation of repetitive text. Additionally, they utilize beam search, which is a heuristic search algorithm that explores the search space in a breadth-first manner to find the most likely output sequence. It extends the search to the top 'B' candidates at each step, where 'B' is a predefined beam width. The beam width determines the number of alternatives (branches) to explore at each step. At each time step, it keeps track of the top 'B' sequences based on the probabilities of the sequences. The final output sequence is the one that has the highest overall score. Beam search offers a good trade-off between the quality of output and computational efficiency. \par
    \begin{figure}[htbp]
      \centering
      \includegraphics[width=12cm]{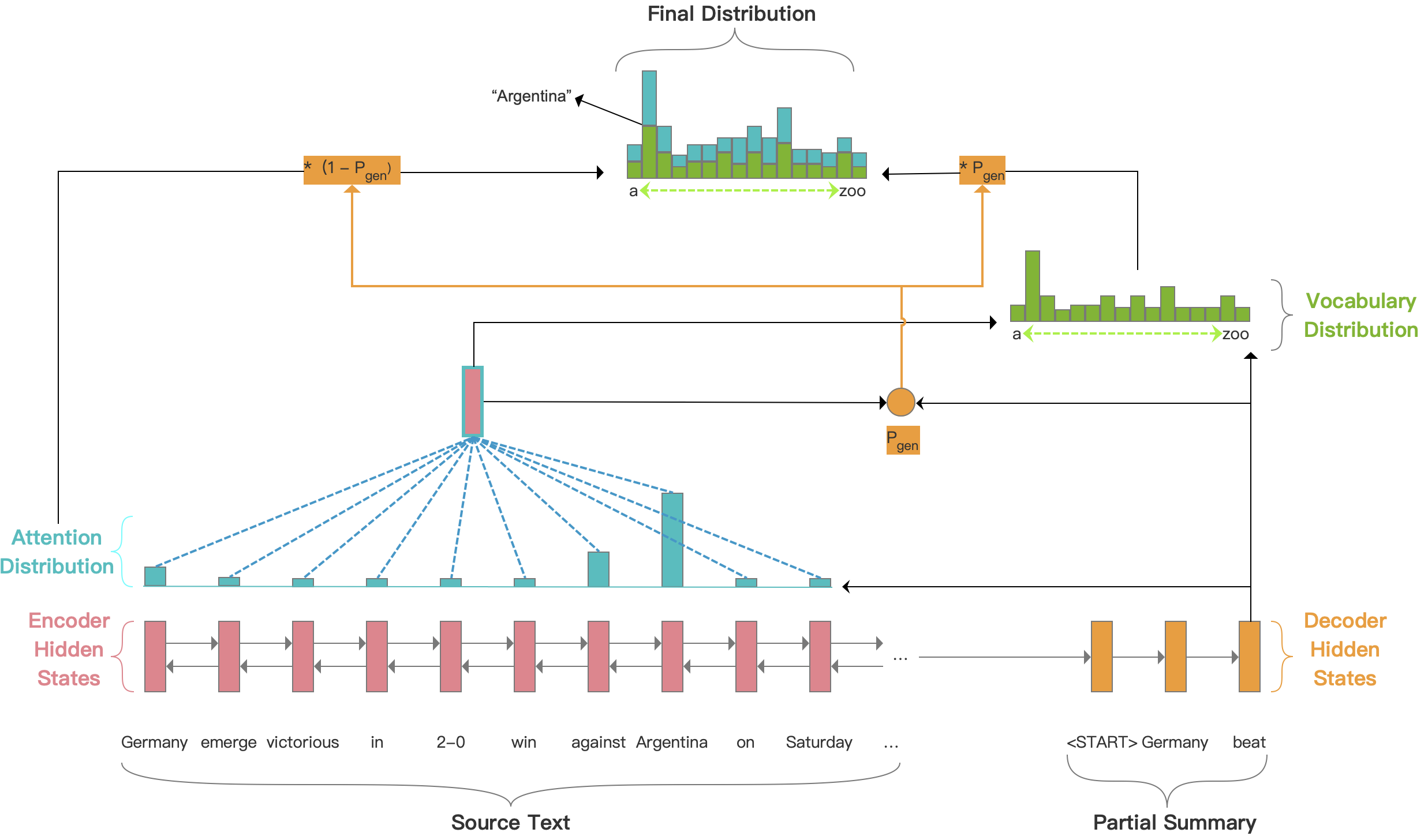}
      \vspace*{8pt}
      \caption{Pointer-generator model combines an RNN with attention and copy network. \cite{see-etal-2017-get}}
    \end{figure}
    Zheng et al. \cite{zheng-etal-2019-subtopic} propose a new method for multi-document summarization called Subtopic-Driven Summarization. The authors argue that each underlying topic of the documents can be seen as a collection of different subtopics of varying importance. The proposed model uses these subtopics and generates the underlying topic representation from a document and subtopic view. The goal is to minimize the difference between the representations learned from the two views. The model uses a hierarchical RNN to encode contextual information and estimates sentence importance hierarchically considering subtopic importance and relative sentence importance. \par
    
    \subsection{Convolutional neural network}
    In NLP, Convolutional neural networks (CNN) \cite{fukushima1980neocognitron, lecun1998gradient, krizhevsky2017imagenet}, which originally developed for image processing and computer vision tasks, are used to capture local patterns and structures within a text, making them particularly effective for tasks that involve extracting meaningful features from a sequence of words or characters. In a typical CNN architecture for NLP, the input text is first represented as a sequence of word or character embeddings, forming a matrix where each row corresponds to a word or character embedding. The embeddings are learned during the training process, allowing the model to capture meaningful representations of words or characters. The main building block of CNN is the convolutional layer, which consists of multiple filters. Each filter is applied to sliding windows of fixed size across the input text, capturing local patterns within the text. The filter's output, or feature map, is then passed through a non-linear activation function, such as the Rectified Linear Unit, to introduce non-linearity into the model. After the convolutional layers, the feature maps are typically passed through a pooling layer, such as max-pooling or average pooling, which reduces the spatial dimensions and extracts the most salient features from the feature maps. This process helps to reduce the computational complexity of the model and makes it more robust to variations in the input. The final layers of a CNN for NLP usually consist of one or more fully connected layers, which combine the extracted features and perform the specific NLP task, such as text classification or sequence tagging. These layers are often followed by a softmax layer for multi-class tasks to produce probability distributions over the possible output classes. \par
    Narayan et al. \cite{narayan-etal-2018-dont} introduce a new concept called "extreme summarization", which aims to generate a single sentence summary that can answer the question "What is the article about?". The researchers proposed a novel abstractive model that is conditioned on the article's topics and based entirely on CNN. This model was found to effectively capture long-range dependencies and identify pertinent content in a document. They also incorporate topic-sensitive embeddings to enhance word context with their topical relevance to the documents. \par
    \begin{figure}[htbp]
      \centering
      \includegraphics[height=8cm]{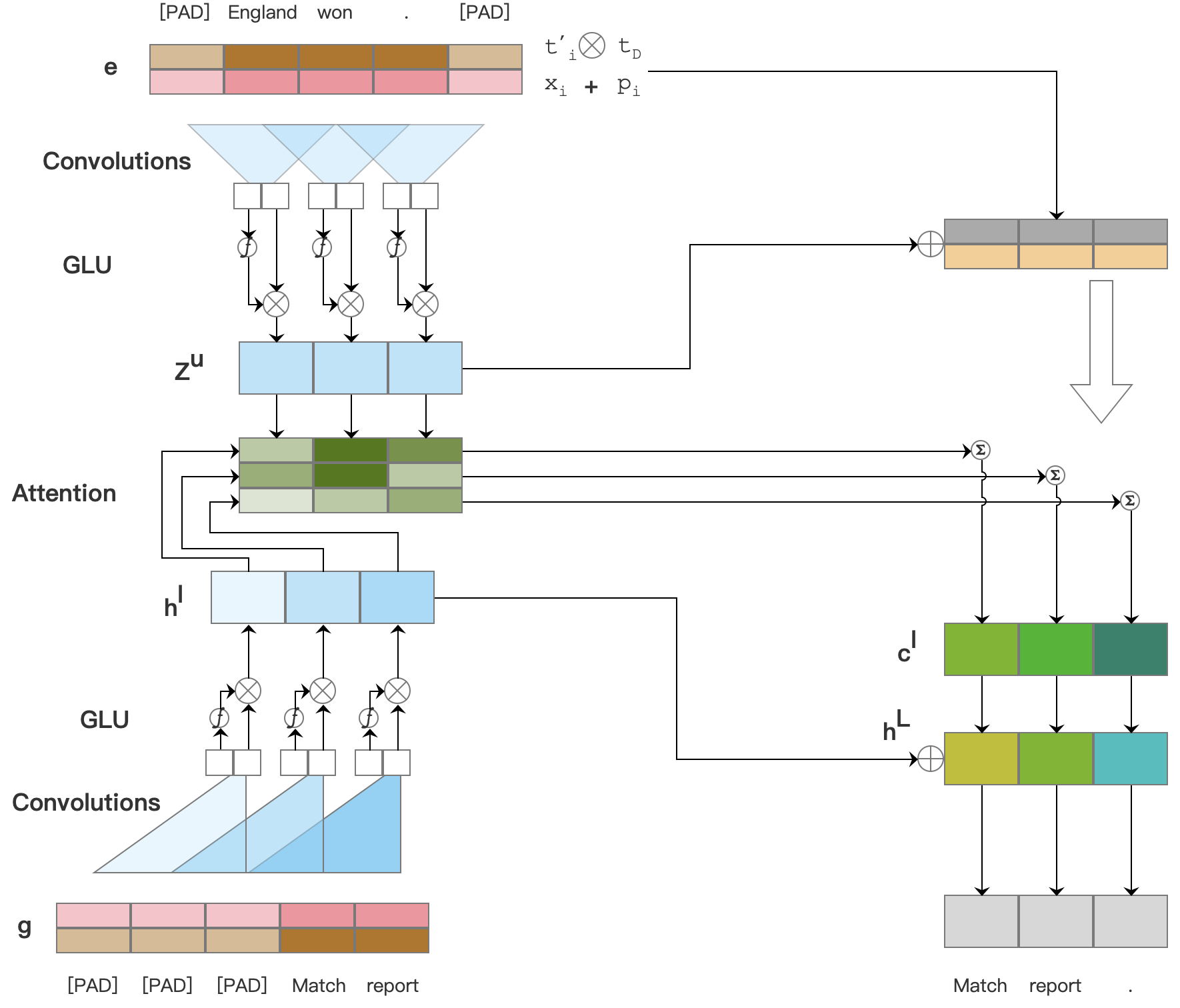}
      \vspace*{8pt}
      \caption{A convolutional model conditioned on the topic for extreme summarization. \cite{narayan-etal-2018-dont}}
    \end{figure}
    Liu et al. \cite{liu-etal-2018-controlling} discuss a new approach for generating summaries of user-defined lengths using CNN. The proposed approach modifies a convolutional sequence-to-sequence model to include a length constraint in each convolutional block of the initial layer of the model. This is achieved by feeding the desired length as a parameter into the decoder during the training phase. At test time, any desired length can be provided to generate a summary of approximately that length. This research contributes a potentially effective method for producing summaries of arbitrary lengths, which holds promise for diverse applications across various tasks requiring summaries of different lengths. \par

    \subsection{Graph neural networks}
    Although Graph Neural Networks (GNN) \cite{scarselli2008graph, micheli2009neural, perozzi2014deepwalk, grover2016node2vec} are primarily used in domains where data naturally exhibits graph structures, such as social networks, molecular structures, and knowledge graphs, they can also be adapted for NLP tasks. In NLP, GNNs are typically used to model relationships between words, sentences, or documents by representing them as nodes in a graph, with edges representing the relationships between these nodes. A GNN model learns to propagate information through the graph by iteratively updating the node representations based on the information from their neighbors. The core building blocks of GNNs are graph convolutional layers, which are designed to aggregate information from neighboring nodes and update the node features.\par
    Jing et al. \cite{jing-etal-2021-multiplex} present a novel Multiplex Graph Convolutional Network (Multi-GCN) approach for extractive summarization. Multi-GCN learns node embedding of different relations among sentences and words separately and combines them to produce a final embedding. This approach helps to mitigate the over-smoothing and vanishing gradient problems of the original GCN. \par
    A heterogeneous GNN, HETERSUMGRAPH \cite{wang-etal-2020-heterogeneous} is introduced for extractive document summarization. This network includes nodes of different granularity levels apart from sentences, which act as intermediaries and enrich cross-sentence relations. This approach allows different sentences to interact considering overlapping word information. Moreover, the graph network can accommodate additional node types, such as document nodes for multi-document summarization. \par
    Doan et al. \cite{doan-etal-2022-multi} propose a method for long document summarization by applying Heterogeneous Graph Neural Networks (HeterGNN) and introducing a homogeneous graph structure (HomoGNN). The HomoGNN focuses on sentence-level nodes to create a graph structure, enriching inter-sentence connections. Simultaneously, the HeterGNN explores the complex relationships between words and sentences, tackling intra-sentence connections. Both networks are constructed and updated using a Graph Attention Network model. In the HomoGNN, a BERT model is used for the initial encoding of sentences, while the HeterGNN uses a combination of CNN and BiLSTM for node feature extraction. After processing, the outputs of both networks are concatenated to form the final representation of sentences. \par

    \begin{figure}[htbp]
      \centering
      \includegraphics[height=8cm]{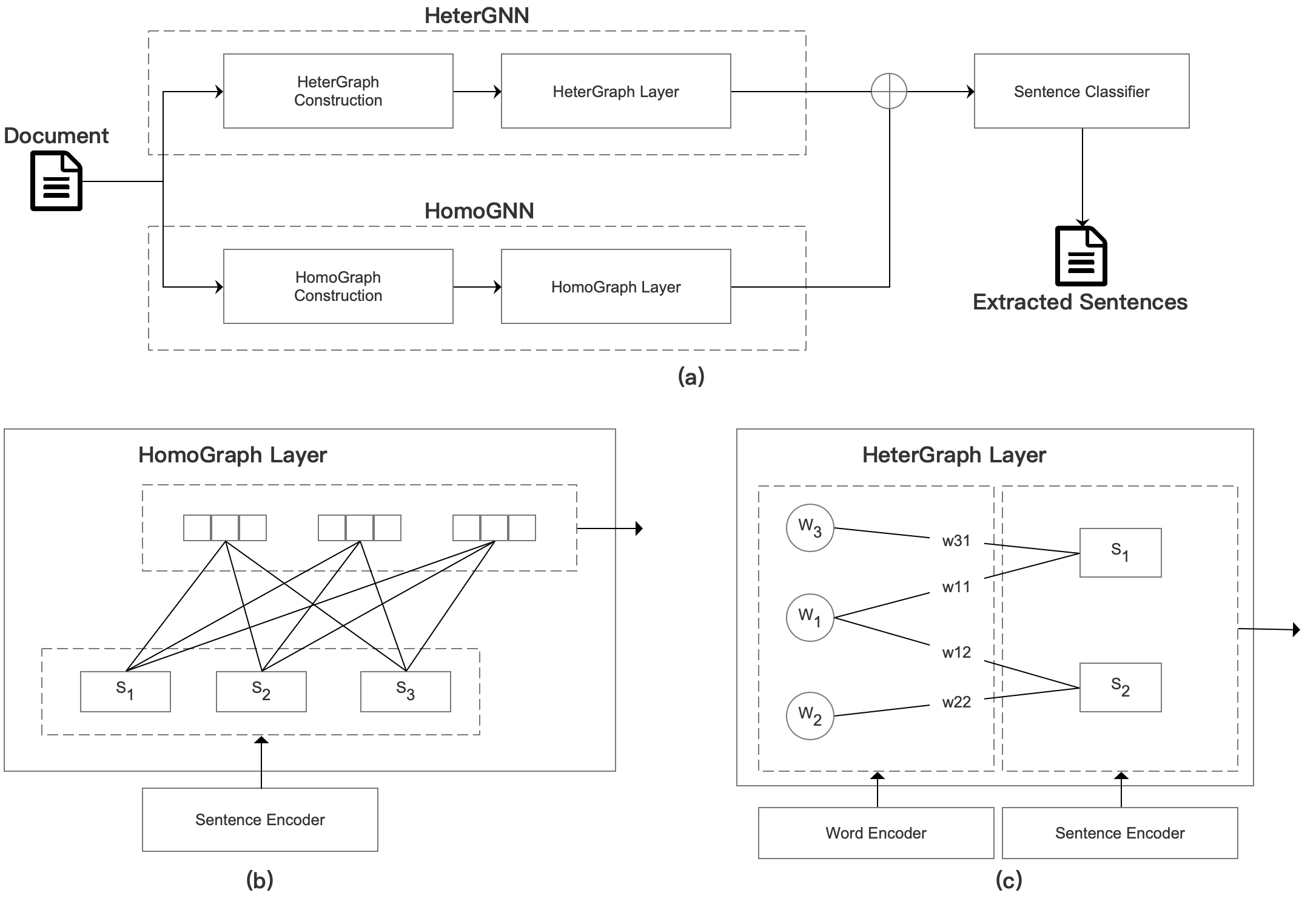}
      \vspace*{8pt}
      \caption{A two-phase pipeline model for extreme summarization \cite{doan-etal-2022-multi}. In the first phase, sentences are encoded using pre-trained BERT, and the [CLS] token information is passed through a graph attention layer. In the second phase, both word and sentence nodes are encoded and fed into a heterogeneous graph layer. The outputs from the two phases are concatenated and inputted into a multi-layer perceptron (MLP) layer for sentence label classification. }
    \end{figure}

    \subsection{Transformer}
    Transformers models \cite{vaswani2017attention, devlin-etal-2019-bert, radford2018improving}, which are the most popular deep learning architecture, have been a revolutionary force in the field of NLP, especially in text summarization. Unlike RNN or LSTM, Transformers use the self-attention mechanism, allowing them to capture dependencies regardless of their distance in the input text. This is particularly useful in text summarization tasks, where understanding the full context of a document is crucial. \par
    Transformer \cite{vaswani2017attention} follows a Seq2Seq architecture and consists of an encoder to map the input sequence into a latent space, and a decoder to map the latent representation back into an output sequence. At the heart of the Transformer model is the self-attention mechanism, which allows the model to weigh the significance of words in the input sequence when generating each word in the output sequence. \par
    BERT \cite{devlin-etal-2019-bert}, which is Bidirectional Encoder Representations Transformers, pre-trains deep bidirectional representations from the unlabeled text by conditioning on both left and right context in all layers. After pre-training, the BERT model can be fine-tuned with an additional output layer to create state-of-the-art models for a wide range of tasks, including summarization, without task-specific architecture modifications. During the fine-tuned phase, the model is initialized with the pre-trained parameters, and all parameters are fine-tuned using labeled data from the downstream tasks. To use BERT for text summarization, a common method is to fine-tune it on a summarization task. BERTSUM \cite{liu2019fine} is an approach to utilize BERT for extractive summarization by adding an interval segment embedding and a positional embedding to the pre-trained BERT model, allowing the model to recognize sentences and their orders. These embeddings are learned during the fine-tuning process. During inference, the most important sentences are selected based on their scores to form the final summary. \par
    \begin{figure}[htbp]
      \centering
      \includegraphics[width=12cm]{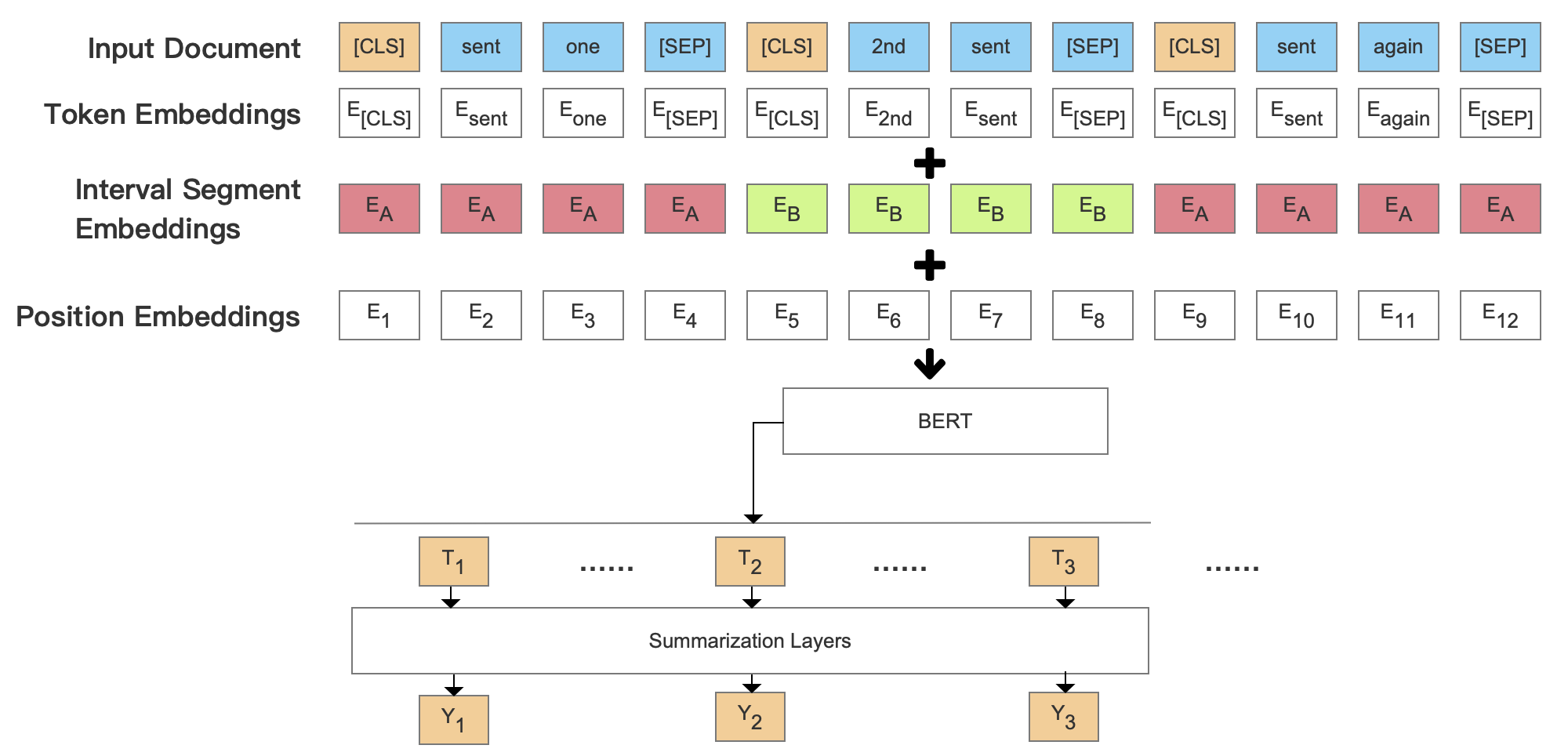}
      \vspace*{8pt}
      \caption{The overview architecture of the BERTSUM model \cite{liu2019fine}.}
    \end{figure}
    T5 \cite{raffel2020exploring}, short for "Text-to-Text Transfer Transformer", is a unified model that treats every NLP problem as a text generation task, enabling the model to learn multiple tasks simultaneously and to learn shared representations across these tasks. In the case of text summarization, the model is trained to predict the summarized text given the original text prefixed with a task-specific instruction, like "summarize:", so the model learns to generate the summary based on the context and the given task. T5 is trained using a denoising auto-encoding objective, which is essentially a causal language modeling task with some noise in the input data. The model has to learn to predict the original clean text from the noisy version. This method forces the model to learn to understand and generate grammatically correct and contextually relevant text, a skill that's very useful in generating coherent and relevant summaries. \par
    BART (Bidirectional and Auto-Regressive Transformers) \cite{lewis-etal-2020-bart}, is a denoising auto-encoder used for pertaining Seq2Seq models. The system works by corrupting text with an arbitrary noising function and then training the model to reconstruct the original text, using a standard Transformer-based neural machine translation architecture. This architecture generalizes the approach used with a bidirectional encoder and a left-to-right decoder and is particularly effective when fine-tuned for text generation tasks like summarization. Several different noising strategies were tested in BART, with the best performance achieved by randomly shuffling sentence order and using an innovative in-filling scheme, where segments of text are replaced with a single mask token. This forces the model to consider overall sentence length and make longer-range transformations to the input. \par
    Pegasus \cite{zhang2020pegasus}, which stands for Pre-training with Extracted gap sentences for Abstractive Summarization, is a model that specifically focuses on abstractive text summarization. Pegasus's main novelty lies in its pre-training strategy, which simulates summarization by masking certain sentences in the document. Instead of masking individual words or phrases, Pegasus masks entire sentences, treating the task as a sentence-level extraction problem. During this phase, the model learns to predict the 'masked' sentences based on the rest of the text, developing a strong sense of sentence-level importance and relevance skills that are vital for text summarization. The model was tested on diverse domains including news, science, stories, and legislative bills, and showed strong performance on all tested datasets, as well as low-resource summarization. \par
    BIGBIRD \cite{zaheer2020big}, a sparse attention mechanism designed to tackle the quadratic dependency of the sequence length, which is a limitation of Transformer-based models like BART. The BIGBIRD mechanism reduces this quadratic dependency to linear, meaning it can handle sequences up to 8 times longer than previously possible on the same hardware. This means that BIGBIRD can understand and generate much longer pieces of text, making it potentially useful for long document summarization. The model consists of three parts: a set of global tokens attending to all parts of the sequence, all tokens attending to a set of local neighboring tokens, and all tokens attending to a set of random tokens. BIGBIRD retains all the known theoretical properties of full transformers and extends the application of the attention-based model to tasks where long contexts are beneficial. \par
    Longformer \cite{beltagy2020longformer} is another modification of the Transformer model designed to handle long sequences. The Longformer offers a linearly scaling attention mechanism to make it possible to process documents with thousands of tokens or more. Longformer's attention mechanism is a combination of local windowed self-attention and task-motivated global attention. This drop-in replacement for the standard self-attention could achieve state-of-the-art results on character-level language modeling tasks. Additionally, Longformer-Encoder-Decoder (LED) is introduced as a variant of Longformer designed for long document generative Seq2Seq tasks. This model is also tested and proven effective on the long document summarization dataset. \par
    \begin{figure}[htbp]
      \centering
      \includegraphics[width=12cm]{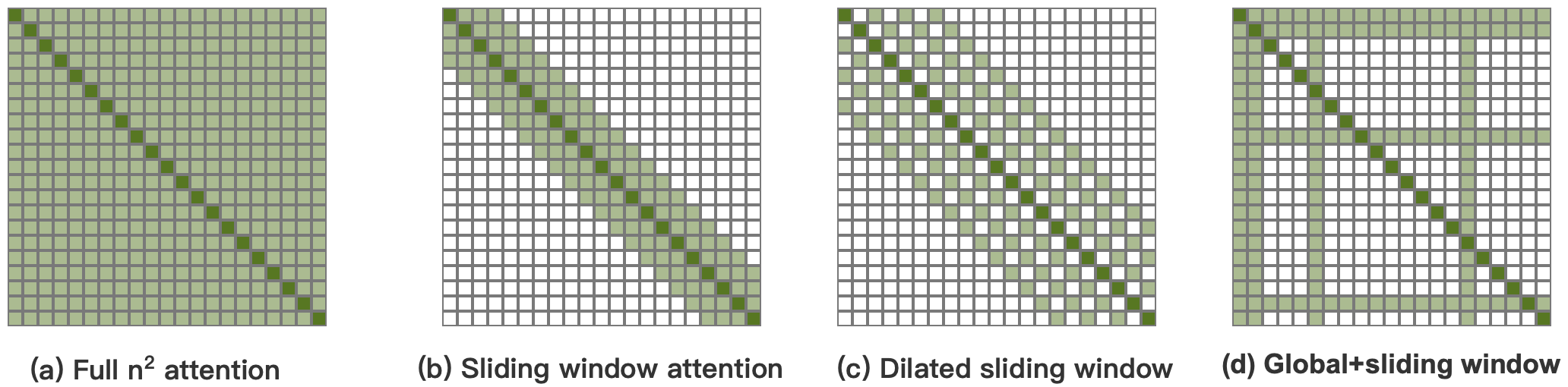}
      \vspace*{8pt}
      \caption{By comparing the complete self-attention pattern and the attention pattern configuration in the Longformer model, we can observe the differences. \cite{beltagy2020longformer}.}
    \end{figure}
    GPT-1 \cite{radford2018improving} (Generative Pre-trained Transformer) is an auto-regressive language model, which can generate human-like by predicting the next word in a sequence. In the context of text summarization, GPT-1 is able to create summaries that are not just extracts of the original text but can rephrase or reframe the content in novel ways, capturing the essence of the document while potentially reducing its length significantly. However, since GPT-1 does not explicitly model the structure of the document beyond the sequence of words, it might not always maintain the coherence and relevance of the summary, especially for long or complex documents. GPT-2 \cite{radford2019language} is an improvement over GPT-1, having more parameters and trained on a larger corpus of data. One important advantage of GPT-2 for summarization is its ability to generate fluent and coherent text due to its training objective and architecture. It can capture long-range dependencies in the text, rephrase the original text, and even generate novel sentences that were not in the original document but are consistent with its content. Furthermore, GPT-2 can be employed for summarization in a few-shot/zero-shot manner where the model is designed to make accurate predictions given only a few or no examples. GPT-3 \cite{brown2020language} follows the design of its predecessor and has been found to generate exceptionally fluent and coherent text based on a larger model size- 175 billion parameters. GPT-3 has a larger context window, meaning it can consider a significant portion of a document when generating a summary. This allows for more holistic and comprehensive summaries, especially when compared to models with smaller context windows that might not capture all necessary information. Apart from the typical summarization tasks, GPT-3 can be leveraged for a range of different summary types. Whether you're interested in producing extractive or abstractive summaries, single-document or multi-document summaries, GPT-3 can be utilized to generate them. InstructGPT \cite{ouyang2022training}, a recent development, centers around training large language models to comprehend and follow instructions with the aid of human feedback. The authors employ a fine-tuning approach on GPT-3, utilizing a dataset of labeler demonstrations that outline the desired model behavior, starting from labeler-written prompts and responses. They initially fine-tuned the model using supervised learning and subsequently employed reinforcement learning techniques with human feedback for further fine-tuning. The authors' findings indicate that fine-tuning models with human feedback hold promise in aligning language models with human intent, highlighting a fruitful direction for future research. Recently, a substantial multi-modal model known as GPT-4 \cite{openai2023gpt4} has emerged. It has the ability to process both image and text inputs and generate text outputs. In human exam evaluations, this model demonstrates exceptional performance, consistently outscoring the majority of human test takers. Despite the lack of specific information regarding the model's structure, hyper-parameters, and training methodology in the paper, the results of GPT-4 exhibit remarkable advancements across diverse tasks, including summarization. Even though GPT-4 might introduce details or points that were not part of the original document, leading to "hallucinations", the development of this model marks a significant milestone in the advancement of AI systems that are both widely applicable and safely deployable. \par
    \begin{figure}[htbp]
      \centering
      \includegraphics[width=12cm]{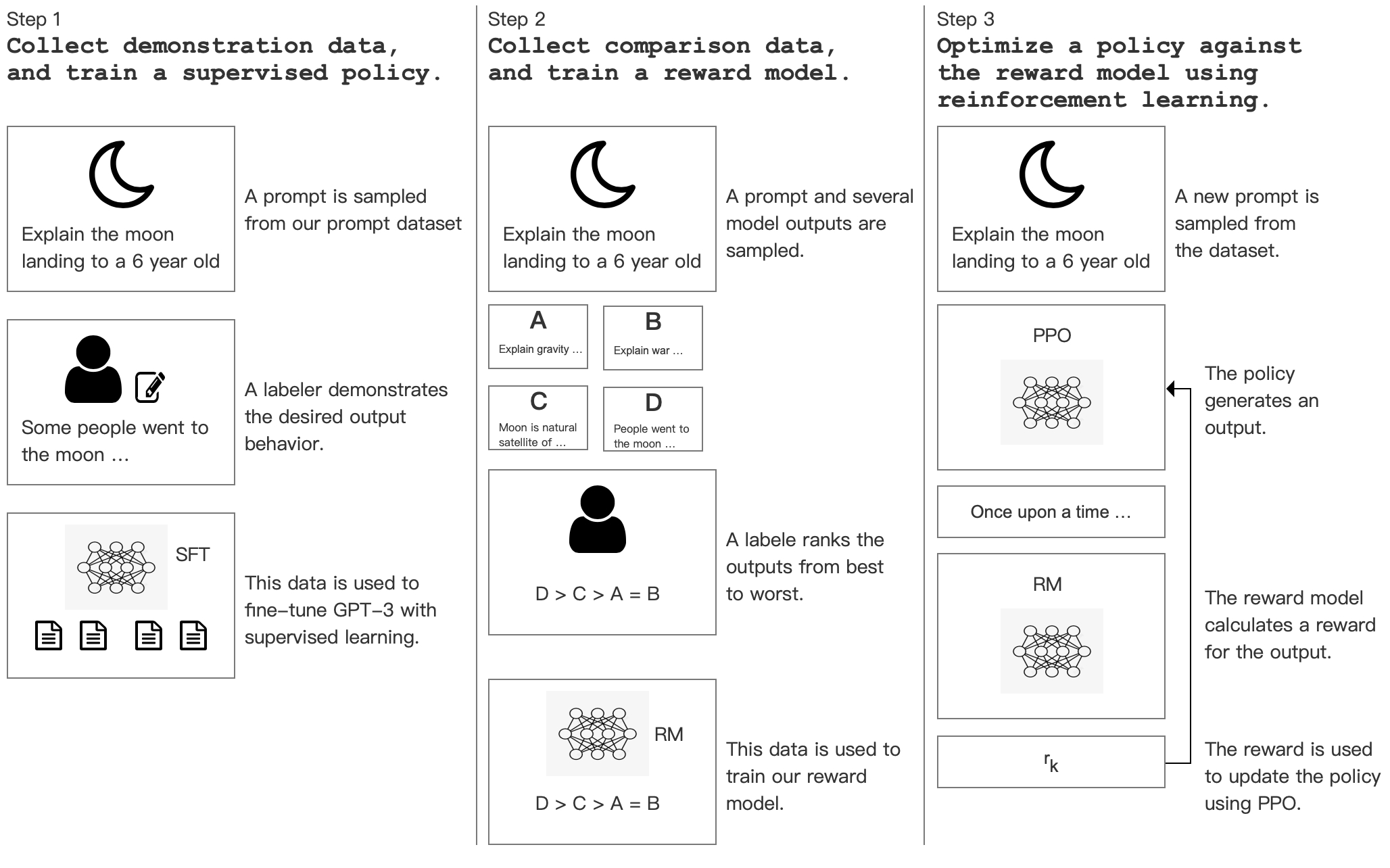}
      \vspace*{8pt}
      \caption{The diagram showcases the three sequential steps of InstructGPT: (1) supervised fine-tuning (SFT), (2) reward model (RM) training, and (3) reinforcement learning through proximal policy optimization (PPO) using this reward model. The blue arrows indicate the utilization of this data to train InstructGPT. \cite{ouyang2022training}.}
    \end{figure}
    \subsection{Reinforcement learning}
    Reinforcement Learning (RL) \cite{kaelbling1996reinforcement, mnih2015human} is a type of Machine Learning where an agent learns to behave in an environment, by performing certain actions and receiving rewards (or punishments) in return. \par
    In the context of text summarization, the environment consists of the input document that needs to be summarized. The state could be the current part of the document being considered for summarization, along with the portion of the summary that has already been generated. The action might involve selecting a sentence from the document to include in the summary (for extractive summarization) or generating a sentence or phrase to add to the summary (for abstractive summarization). The reward is a measure of the quality of the generated summary. This could be based on a variety of factors, such as how well the summary represents the main points of the document, how grammatically correct and fluent it is, and so on. \par
    One of the key benefits of using RL for text summarization is that it allows for a more flexible and adaptive approach to summarization, compared to traditional supervised learning methods. RL can learn to adapt its summarization strategy based on the specific characteristics of each document and can optimize for long-term rewards (like the overall coherence and quality of the summary), rather than just short-term gains (like the accuracy of the next sentence). \par
    Narayan et al. \cite{narayan-etal-2018-ranking} proposes a novel method for single document summarization using extractive summarization as a sentence ranking task, globally optimizing the ROUGE evaluation metric through a reinforcement learning objective. The authors argue that current cross-entropy training is sub-optimal for extractive summarization, tending to generate verbose summaries with unnecessary information. Their method improves this by learning to rank sentences for summary generation. The approach involves viewing the neural summarization model as an "agent" in a reinforcement learning paradigm, which reads a document and predicts a relevance score for each sentence. The agent is then rewarded based on how well the extract resembles the gold-standard summary. The REINFORCE algorithm is used to update the agent, optimizing the final evaluation metric directly instead of maximizing the likelihood of the ground-truth labels, making the model more capable of discriminating among sentences. \par
    \begin{figure}[htbp]
      \centering
      \includegraphics[width=12cm]{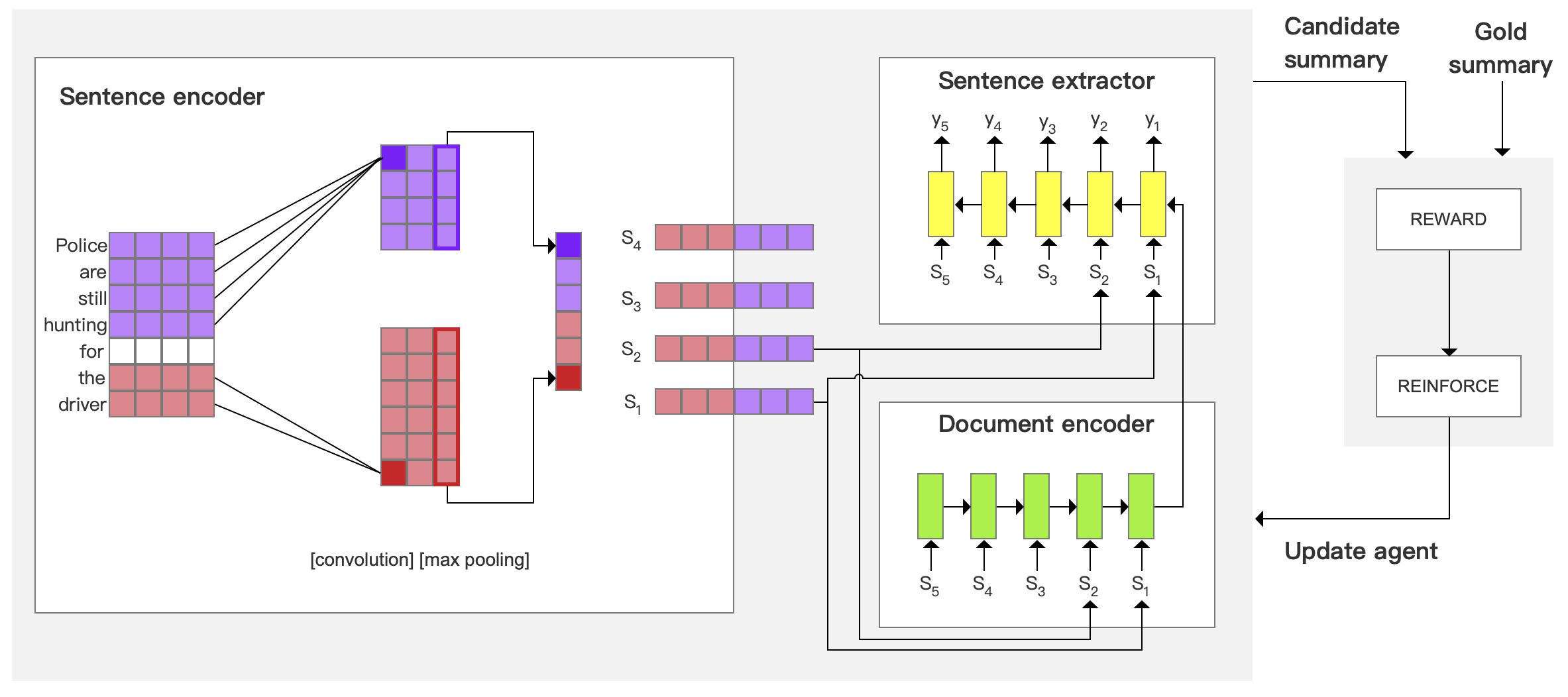}
      \vspace*{8pt}
      \caption{The extractive summarization model with reinforcement learning \cite{narayan-etal-2018-ranking} employs a hierarchical encoder-decoder architecture to rank sentences based on their extract worthiness. A candidate summary is then formed by assembling the top-ranked sentences. The REWARD generator assesses the candidate summary against the gold summary, providing a reward signal that is utilized in the REINFORCE algorithm \cite{williams1992simple} to update the model.}
    \end{figure}
    Hyun et al. \cite{hyun-etal-2022-generating} introduce another model for unsupervised abstractive sentence summarization using reinforcement learning (RL). Unlike previous methods, which mainly utilize extractive summarization (removing words from texts), this method is abstractive, allowing for the generation of new words not found in the original text, thereby increasing flexibility and versatility. The approach involves developing a multi-summary learning mechanism that creates multiple summaries of varying lengths from a given text, with these summaries enhancing each other. The RL-based model used assesses the quality of summaries using rewards, considering factors such as the semantic similarity between the summary and the original text and the fluency of the generated summaries. The model also involves a pre-training task to achieve well-initialized model parameters for RL training. \par
    
\section{Data and Experimental Performance Analysis}
    \subsection{Techniques for data processing}
    This sub-section will delve into the key techniques utilized for data processing.
    \subsubsection{Pre-training}
    In the domain of text summarization, pre-training refers to the initial training phase of a model on a large, diverse corpus of text data, prior to fine-tuning it on a more specific summarization task. This strategy capitalized on the capabilities of extensive language models such as GPT \cite{radford2018improving}, BART \cite{lewis-etal-2020-bart}, and T5 \cite{raffel2020exploring}, which have been pre-trained on vast quantities of text data to comprehend syntactic and semantic patterns within a language. \par
    \subsubsection{Few-shot, zero-shot learning}
    Few-shot and zero-shot learning are terms used to describe scenarios where a model is required to make accurate predictions for new classes that were either minimally represented (few-shot) or completely absent (zero-shot) during the training phase \cite{xian2018zero, sung2018learning}. Few-shot learning in summarization usually implies a scenario where the model is trained on many examples from a few categories and is then expected to generalize to summarizing examples from new categories after seeing only a few examples from these new categories. Zero-shot learning in summarization, on the other hand, refers to a scenario where the model is expected to generalize to entirely new categories without seeing any examples from these categories during training. The idea behind these methods is to provide the model with a few examples or a description of the task at inference time, allowing it to adjust its predictions based on this new information. This often involves formulating the summarization task as a type of prompt that the model is designed to complete. \par
    \subsubsection{Prompting}
    Prompts play a crucial role in the current generation of language models, particularly those that are trained in a transformer-based architecture like GPT-4 \cite{openai2023gpt4} or T5 \cite{raffel2020exploring}. The term "prompt" in the context of these models refers to the input given to the model to indicate the task it should perform. For text summarization, the prompt is typically the text that needs to be summarized. However, in the case of GPT-4 \cite{openai2023gpt4}, T5 \cite{raffel2020exploring}, and similar models, the prompt can also include a task description or examples to guide the model's generation. This is especially important in few-shot and zero-shot learning scenarios. Choosing effective prompts is a bit of an art and can significantly impact the performance of the model. The best practices for creating prompts are still an active area of research, but a well-designed prompt often includes clear instructions and, when possible, an example of the desired output. \par
    \begin{figure}[htbp]
      \centering
      \includegraphics[width=12cm]{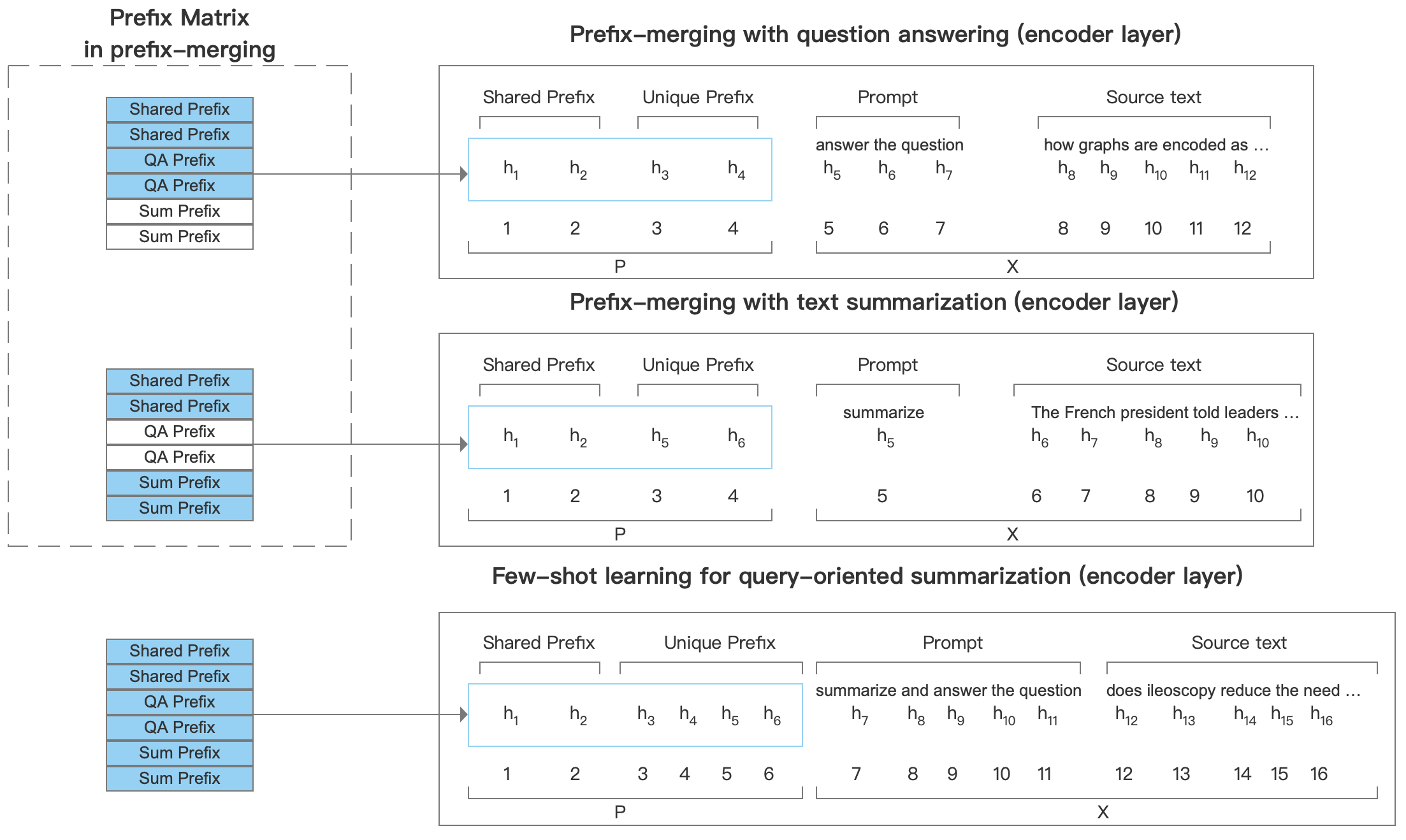}
      \vspace*{8pt}
      \caption{The figure highlights the encoder layer of BART and provides annotated examples and a comparison between prefix-merging on the two auxiliary tasks (top, mid) and the application of the merged prefix on the Few-shot Query-Focused Summarization task using prefix-tuning (bottom). \cite{yuan-etal-2022-shot}}
    \end{figure}
    \subsubsection{Domain adaptation}
    Domain adaptation in the context of text summarization pertains to the process of adapting a summarization model, initially trained on a specific domain (e.g., news articles), to perform effectively on a different yet related domain (e.g., scientific papers or legal documents) \cite{mao-etal-2022-citesum, moro2022semantic}. Fine-tuning the model on a smaller dataset from the target domain is a commonly employed approach. Additionally, transfer learning is utilized to capitalize on the knowledge acquired from one domain and apply it to another. \par
    \subsubsection{Tokenization, embedding and Decoding strategies}
    Tokenization is the initial step of dividing the text into individual units known as tokens, which serve as fundamental input elements for most natural language processing (NLP) models \cite{webster1992tokenization, manning-etal-2014-stanford}. Following tokenization, the tokens are converted into continuous vector representations through an embedding layer to create the input embeddings \cite{pennington-etal-2014-glove}. These embeddings are then fed into the model for further processing. Once the model generates an output sequence, such as a summary, the reverse process takes place, where each token is mapped back to its corresponding word in a vocabulary, and the words are subsequently joined together using spaces to produce human-readable text. \par
    \subsection{Dataset and Evaluation Metrics}
    In the field of text summarization, various datasets and evaluation metrics are used to train models and assess their performance. Each evaluation metric has its strengths and weaknesses. While some metrics are easy to compute, they might not accurately reflect the human judgment of quality as they are based solely on n-gram overlap. Some other metrics attempt to address this issue by taking semantic similarity into account. However, manual evaluation by human judges is still considered the gold standard, despite its scalability challenges. We will first discuss some commonly used datasets, and then we'll talk about evaluation metrics. \par
    \subsubsection{Dataset}
    Table \ref{table2} presents a compilation of notable datasets in the field of text summarization. The "Size" column indicates the respective counts of training, validation, and test documents available in each dataset. \par
    \begin{center}
    \begin{table}
        \tbl{\label{table2}A comparison of different Datasets}{
        \begin{tabular}{cccccc}
        \toprule
        Dataset & Domain & Tasks & Size & \makecell{Document \\ Length} & \makecell{Summary \\ Length}\\
        \hline
        DUC2004 \cite{DUC2004} & News & \makecell{single-document \\ multi-document \\ cross-lingual \\ query-focused \\ Abstractive} & 500 & -  & \makecell{$<$= 75 bytes (single) \\ $<$= 665 bytes (multi)}\\
        \hline
        \makecell{CNN/Daily \\Mail \cite{nallapati-etal-2016-abstractive}} & News & \makecell{single-document \\ Headline \\ query-focused \\ Abstractive \\ Extractive} & \makecell{286,817 \\ 13,368 \\ 11,487} & \makecell{766 words \\ average} & \makecell{53 words \\ average} \\
        \hline
        XSum \cite{narayan-etal-2018-dont} & News & \makecell{single-document \\ cross-lingual \\ Abstractive} & \makecell{204,045 \\ 11,332 \\ 11,334} & \makecell{431 words \\ average} & \makecell{23 words \\ average} \\
        \hline
        WikiSum \cite{liu2018generating} & Wiki & \makecell{multi-document \\ Abstractive} & \makecell{1,865,750 \\ 233,252 \\ 232,998} & \makecell{-} & \makecell{-} \\
        \hline
        Multi-News \cite{fabbri-etal-2019-multi} & News & \makecell{multi-document \\ Abstractive \\ Extractive}  & \makecell{ 44,972 \\  5,622 \\ 5,622} & \makecell{2,103 words \\ average} & \makecell{263  words \\ average} \\
        \hline
        BillSum \cite{kornilova-eidelman-2019-billsum} & Legal & \makecell{single-document \\ Long-document \\ Abstractive} & \makecell{18,949 \\ 1,237 \\ 3,269} & \makecell{1,592 words \\ average} & \makecell{197 words \\ average} \\
        \hline
        PubMed \cite{cohan-etal-2018-discourse} & Medical & \makecell{single-document \\ Long-document \\ Abstractive} & \makecell{119,924 \\ 6,633 \\ 6,658} & \makecell{3,016 words \\ average} & \makecell{203 words \\ average} \\
        \hline
        arXiv \cite{cohan-etal-2018-discourse} & Scientific & \makecell{single-document \\ Long-document \\ Abstractive} & \makecell{203,037 \\ 6,436 \\ 6,440} & \makecell{4,938 words \\ average} & \makecell{220 words \\ average} \\
        \hline
        XGLUE \cite{liang-etal-2020-xglue} & News & \makecell{cross-lingual \\ Headline \\ Abstractive} & \makecell{300,000 \\ 50,000 \\ 50,000} & \makecell{-} & \makecell{-} \\
        \hline
        BIGPATENT \cite{sharma-etal-2019-bigpatent} & Patent & \makecell{single-document \\ Long-document \\ Extractive \\ Abstractive} & \makecell{1,207,222 \\ 67,068 \\ 67,072} & \makecell{3,572 words \\ average} & \makecell{116 words \\ average} \\
        \hline
        Newsroom \cite{grusky-etal-2018-newsroom} & News & \makecell{single-document \\Headline \\ Extractive \\ Abstractive} & \makecell{995,041 \\ 108,837 \\ 108,862} & \makecell{658 words \\ average} & \makecell{26 words \\ average} \\
        \hline
        MLSUM \cite{scialom-etal-2020-mlsum} & News & \makecell{single-document \\ cross-lingual \\ Multi-lingual \\ Abstractive} & \makecell{287,096 \\ 11,400 \\ 10,700} & \makecell{790 words \\ average} & \makecell{55 words \\ average} \\
        \botrule
        \end{tabular}}
        \end{table}
    \end{center}
    
    \subsubsection{Evaluation Metrics}
    Evaluating the quality of generated summaries is crucial in the text summarization task. Here is an assortment of evaluation metrics commonly employed in summarization:
    \begin{itemize}
        \item [$\blacktriangleright$] Rouge (Recall-Oriented Understudy for Gisting Evaluation) \cite{lin-2004-rouge}: is a set of evaluation metrics used for evaluating automatic summarization. It compares the system-generated output with a set of reference summaries. ROUGE-N measures the overlap of N-grams (a contiguous sequence of N items from a given sample of text or speech) between the system and reference summaries. It includes metrics like ROUGE-1 (for unigrams), ROUGE-2 (for bigrams), and so on. ROUGE-L metric measures the Longest Common Subsequence (LCS) between the system and reference summaries. LCS takes into account sentence-level structure similarity naturally and identifies the longest co-occurring in sequence n-grams automatically.
        \item [$\blacktriangleright$] BLEU (Bilingual Evaluation Understudy) \cite{papineni-etal-2002-bleu}: is an evaluation metric initially developed for assessing the quality of machine-translated text, but it has also been used in text summarization tasks. It is a precision-based metric that compares the system-generated summary with one or more reference summaries. BLEU operates at the n-gram level to measure the overlap of n-grams between the generated output and the reference texts. It calculates the precision for each n-gram size (usually from 1-gram to 4-gram) and takes a weighted geometric mean to compute the final score.
        \item [$\blacktriangleright$] METEOR (Metric for Evaluation of Translation with Explicit Ordering) \cite{banerjee-lavie-2005-meteor}: is an evaluation metric initially designed for machine translation tasks but also used in text summarization evaluations. Unlike the previously mentioned metrics like ROUGE and BLEU that mainly focus on recall and precision at the n-gram level, METEOR incorporates more linguistic features and tries to align the generated text and the reference at the semantic level, thus potentially capturing the quality of the output more accurately.
        \item [$\blacktriangleright$] Pyramid Score \cite{nenkova-passonneau-2004-evaluating}: is based on the principle that a perfect summary could include any of several valid points from the source text, and as such, it would not be fair to penalize a summary for not including specific points. In the Pyramid Score method, human assessors identify Summary Content Units (SCUs) in a set of model summaries, which are essentially nuggets of information. Each SCU is assigned a weight based on how many model summaries it appears in. The Pyramid Score is then computed for a system-generated summary by adding up the weights of the SCUs it contains and normalizing this sum by the maximum possible score achievable by any summary of the same length. Pyramid scoring acknowledges the potential variation in content selection across different acceptable summaries. However, this method is quite labor-intensive because it requires human assessors to perform detailed content analysis on the model summaries.
        \item [$\blacktriangleright$] CIDEr (Consensus-based Image Description Evaluation) \cite{vedantam2015cider}: is an evaluation metric primarily designed for assessing the quality of image captions in the context of image captioning tasks. It also gets used in text summarization to some extent. The fundamental idea behind CIDEr is that words that are more important to a description should have higher weights.
        \item [$\blacktriangleright$] BERTScore \cite{zhang2019bertscore}: is an automatic evaluation metric for natural language generation tasks, including text summarization. Unlike traditional metrics like ROUGE and BLEU which rely on n-gram overlaps, BERTScore leverages the contextual embeddings from the pre-trained BERT model to evaluate the generated text.
        \item [$\blacktriangleright$] Moverscore \cite{zhao-etal-2019-moverscore}: is based on two fundamental principles: the use of contextualized embeddings and the Earth Mover's Distance (EMD), also known as the Wasserstein distance. Contextualized embeddings, such as BERT embeddings, represent words or phrases within the context they appear, providing a more meaningful representation of the text. The Earth Mover's Distance is a measure of the distance between two probability distributions over a region, and it's used here to measure the distance between the embeddings of the generated summary and the reference summary. 
    \end{itemize}
     Table \ref{table3} shows some experimental results of popular models on CNN/Daily Mail. Although automatic metrics are widely used, they do not always align well with human judgments of summary quality. Human evaluation is considered the gold standard, but it's time-consuming and costly. Thus, a combination of automatic and human evaluation is often used in practice. \par
        \begin{center}
        \begin{table}
        \tbl{\label{table3}A comparison of different Models on CNN/Daily Mail \cite{nallapati-etal-2016-abstractive}}{
        \begin{tabular}{ccccc}
        \toprule
        Models & Model & Rouge-1 & Rouge-2 & Rouge-L\\
        \hline
        Attentional RNN \cite{nallapati-etal-2016-abstractive} & RNN & 35.46 & 13.30 & 32.65 \\
        \hline
        Pointer-Generator \cite{see-etal-2017-get} & RNN & 39.53 & 17.28 & 36.38 \\
        \hline
        DynamicConv \cite{wu2019pay} & CNN & 39.84 & 16.25 & 36.73 \\
        \hline
        TaLK Convolution \cite{lioutas2020time} & CNN & 40.59 & 18.97 & 36.81 \\
        \hline
        RL with intra-attention \cite{paulus2017deep} & RL & 41.16 & 15.75 & 39.08 \\
        \hline
        \makecell{RNN-ext+abs+RL+rerank \cite{chen-bansal-2018-fast}} & \makecell{RNN+RL} & 39.66 & 15.85 & 37.34 \\
        \hline
        \makecell{BILSTM+GNN+LSTM+POINTER \cite{fernandes2021structured}} & \makecell{GNN+LSTM} & 38.10 & 16.10 & 33.20 \\
        \hline
        Graph-Based Attentional LSTM \cite{tan-etal-2017-abstractive} & \makecell{GNN+LSTM} & 38.10 & 13.90 & 34.00\\
        \hline
        Transformer \cite{vaswani2017attention} & Transformer & 39.50 & 16.06 & 36.63 \\
        \hline
        PEGASUS \cite{zhang2020pegasus} & Transformer & 44.17 & 21.47 & 41.11 \\
        \hline
        BART \cite{lewis-etal-2020-bart} & Transformer & 44.16 & 21.28 & 40.90 \\
        \hline
        SEASON \cite{wang-etal-2022-salience} & Transformer & 46.38 & 22.83 & 43.18 \\
        \hline
        BART.GPT-4 \cite{liu2023learning} & Transformer & 63.22 & 44.70 & - \\
        \botrule
        \end{tabular}}
        \end{table}
        \end{center}

\section{Summary}
    \subsection{Challenge and Future}
    Text summarization is an intriguing and challenging task in the realm of natural language processing. In recent years, significant advancements have been made with the aid of deep learning (DL) models. Novel concepts such as neural embedding, attention mechanism, self-attention, Transformer, BERT, and GPT-4 have propelled the field forward, resulting in rapid progress over the past decade. However, despite these advancements, there are still notable challenges that need to be addressed.
    This section aims to highlight some of the remaining challenges in text summarization and explore potential research directions that can contribute to further advancements in the field. By addressing these challenges and exploring new avenues, we can continue to push the boundaries of text summarization and unlock its full potential. \par
    
    One critical aspect is the need to understand the context of the document, encompassing semantics, syntactic structure, and discourse organization. Deep learning models often struggle with complex or ambiguous language, idiomatic expressions, and domain-specific jargon, making it difficult to achieve accurate and meaningful summaries. \par
    A well-crafted summary should exhibit coherence and cohesion, ensuring that ideas logically connect and the text flows smoothly. Models must generate summaries that preserve the integrity of the original text's meaning without introducing inconsistencies or redundancies. This requires a deep understanding of the main ideas, supporting details, and their interrelationships. Identifying important content poses a significant challenge, as it necessitates discerning the relevance and significance of various elements in the document. \par
    Summarizing long documents, such as legal or research papers, presents additional hurdles. These documents often contain complex sentence structures, advanced vocabulary, and important information distributed throughout the text. Creating concise summaries that capture the key points while maintaining accuracy becomes a daunting task. Furthermore, the lack of labeled training data exacerbates the challenges. Supervised learning approaches for text summarization rely on substantial amounts of annotated data, which can be expensive and time-consuming to create. \par
    Evaluating the quality of generated summaries is another ongoing challenge. Automatic evaluation metrics, such as ROUGE, BLEU, or BERTScore, do not always align perfectly with human judgment. Manual evaluation, while providing more accurate insights, is a labor-intensive process. Overcoming these challenges requires the development of better evaluation metrics that align more closely with human perceptions of summary quality. \par
    Summarization tasks become more intricate when they are domain-specific, such as in medical or legal contexts. These domains often employ specialized language, requiring a higher level of understanding and accuracy. Additionally, summarizing information from multiple documents introduces further complexities. Models must eliminate redundant information, handle potentially conflicting details, and synthesize the most relevant content from various sources. \par
    As the field progresses, advancements in deep learning models, particularly transformer-based architectures like BERT, GPT-4, and T5, offer promising opportunities for improved performance in text summarization tasks. Fine-tuning pre-trained models on specific summarization objectives has shown great potential and is expected to continue. Furthermore, as we become increasingly interconnected globally, there will be a growing demand for models capable of summarizing text in different languages or even across languages. \par
    Addressing the challenges of explainability, transparency, bias, data efficiency, multi-modal summarization, and personalized summarization are areas that will likely receive significant attention in future research. Explainable and transparent AI models are becoming increasingly important, and efforts to develop models that can provide reasoning for their decisions are expected. The development of better evaluation metrics, mitigating biases, exploring more data-efficient methods, handling multi-modal information, and catering to personalized summarization needs are all potential avenues for further advancement in the field. \par

    \subsection{Conclusion}
    This article presents a comprehensive survey of over 100 deep learning models developed in the last decade, highlighting their significant advancements in various text summarization tasks. Additionally, we provide an overview of popular summarization datasets and conduct a quantitative analysis to assess the performance of these models on several public benchmarks. Furthermore, we address open challenges in the field and propose potential future research directions.

\bibliographystyle{plain}
\bibliography{main}

\end{document}